\documentclass[10pt,journal,compsoc]{IEEEtran}
\usepackage{times}
\usepackage{epsfig}
\usepackage{graphicx}
\usepackage{amssymb}

\usepackage{amsfonts}
\usepackage{bbm}
\usepackage{booktabs}
\usepackage{algorithm}
\usepackage{subfigure}
\usepackage{multirow}

\usepackage{color}
\usepackage{xcolor}

\ifCLASSOPTIONcompsoc
  \usepackage[nocompress]{cite}
\else
  \usepackage{cite}
\fi
\ifCLASSINFOpdf
\else
\fi
\usepackage{amsmath}
\usepackage{algorithmic}
\begin{document}
%
\title{Binary Graph Convolutional Network with Capacity Exploration}
%
%
%
%

\author{Junfu~Wang,
        Yuanfang~Guo,~\IEEEmembership{Senior~Member,~IEEE,}\\
        Liang~Yang,~Yunhong~Wang,~\IEEEmembership{Fellow,~IEEE}

\IEEEcompsocitemizethanks{
  \IEEEcompsocthanksitem J. Wang, Y. Guo, and Y. Wang are with the School of Computer Science and Engineering, Beihang University, Beijing 100191, China. \\E-mail: \{wangjunfu, andyguo, yhwang\}@buaa.edu.cn.
  \IEEEcompsocthanksitem L. Yang is with the School of Artificial Intelligence, Hebei University of Technology, Tianjin 300401, China. E-mail: yangliang@vip.qq.com. 
  }
}
\markboth{IEEE TRANSACTIONS ON PATTERN ANALYSIS AND MACHINE INTELLIGENCE. Under Review}
{Shell \MakeLowercase{\textit{et al.}}: Bare Demo of IEEEtran.cls for Computer Society Journals}
%



\IEEEtitleabstractindextext{%
\begin{abstract}
  The current success of Graph Neural Networks (GNNs) usually relies on loading the entire attributed graph for processing, which may not be satisfied with limited memory resources, especially when the attributed graph is large.
  This paper pioneers to propose a Binary Graph Convolutional Network (Bi-GCN), which binarizes both the network parameters and input node attributes and exploits binary operations instead of floating-point matrix multiplications for network compression and acceleration.
  Meanwhile, we also propose a new gradient approximation based back-propagation method to properly train our Bi-GCN.
  According to the theoretical analysis, our Bi-GCN can reduce the memory consumption by an average of $\thicksim$31x for both the network parameters and input data, and accelerate the inference speed by an average of $\thicksim$51x, on three citation networks, i.e., Cora, PubMed, and CiteSeer.
  Besides, we introduce a general approach to generalize our binarization method to other variants of GNNs, and achieve similar efficiencies. 
  Although the proposed Bi-GCN and Bi-GNNs are simple yet efficient, these compressed networks may also possess a potential \textit{capacity problem}, i.e., they may not have enough storage capacity to learn adequate representations for specific tasks.
  To tackle this \textit{capacity problem}, an Entropy Cover Hypothesis is proposed to predict the lower bound of the width of Bi-GNN hidden layers.
  Extensive experiments have demonstrated that our Bi-GCN and Bi-GNNs can give comparable performances to the corresponding full-precision baselines on seven node classification datasets and verified the effectiveness of our Entropy Cover Hypothesis for solving the \textit{capacity problem}.
\end{abstract}

}

\maketitle

\IEEEdisplaynontitleabstractindextext

%
\IEEEpeerreviewmaketitle

\IEEEraisesectionheading{\section{Introduction}\label{sec:introduction}}

%
%
%
%

\IEEEPARstart{G}{raph}, which represents data with complicated relationships, is extensively employed in real world applications.
In recent years, Graph Neural Networks (GNNs) have achieved impressive performances in a wide variety of tasks, due to their superior representation abilities on irregular graph data.
Examples include biology prediction \cite{gnn_app_biology1,gnn_app_biology2}, social analysis \cite{gnn_app_social1,gnn_app_social2}, traffic prediction \cite{gnn_app_traffic1,gnn_app_traffic2}, etc.

Unfortunately, current GNNs are designed under an implicit assumption that the input of GNNs contains the entire attributed graph \cite{gcn,maskedgcn,gin}.
If the entire graph is too large to be fed into GNNs (due to limited memory resources), in both the training and inference process, which is highly likely when the scale of the graph increases, the performances of GNNs may degrade drastically.

To tackle this problem, an intuitive solution is sampling, i.e., sampling subgraphs with suitable sizes to be separately loaded into GNNs.
Then, GNNs can be trained with subgraphs via a mini-batch scheme.
The sampling-based methods can be classified into three categories, neighbor sampling \cite{graphsage,pinsage,stgcn}, layer sampling \cite{fastgcn, LADIES}, and graph sampling \cite{zeng2019accurate,cluster-gcn, graphsaint}.
Neighbor sampling selects a fixed number of neighbors for each node in the next layer to ensure that every node can be sampled.
Thus, it can be utilized in both the training and inference process.
Unfortunately, when the number of layers increases, the problem of \textit{neighbor explosion} \cite{graphsaint} arises, such that both the training and inference time will increase exponentially.
Different from neighbor sampling, layer sampling usually samples a subgraph in each layer, while graph sampling constructs a set of subgraphs and builds a full GNN on each subgraph. 
These two types of approaches directly sample subgraphs in the training process thus they can avoid the problem of \textit{neighbor explosion}.
However, they cannot guarantee that every node can be sampled for at least once in the whole training/inference process.
Thus, they are only feasible for the training process, because the testing process usually requires GNNs to process every node in the graph.

\begin{figure}[t]
  \centering
  \subfigure[] { 
    \label{fig1:a}
    \includegraphics[width=0.45\columnwidth]{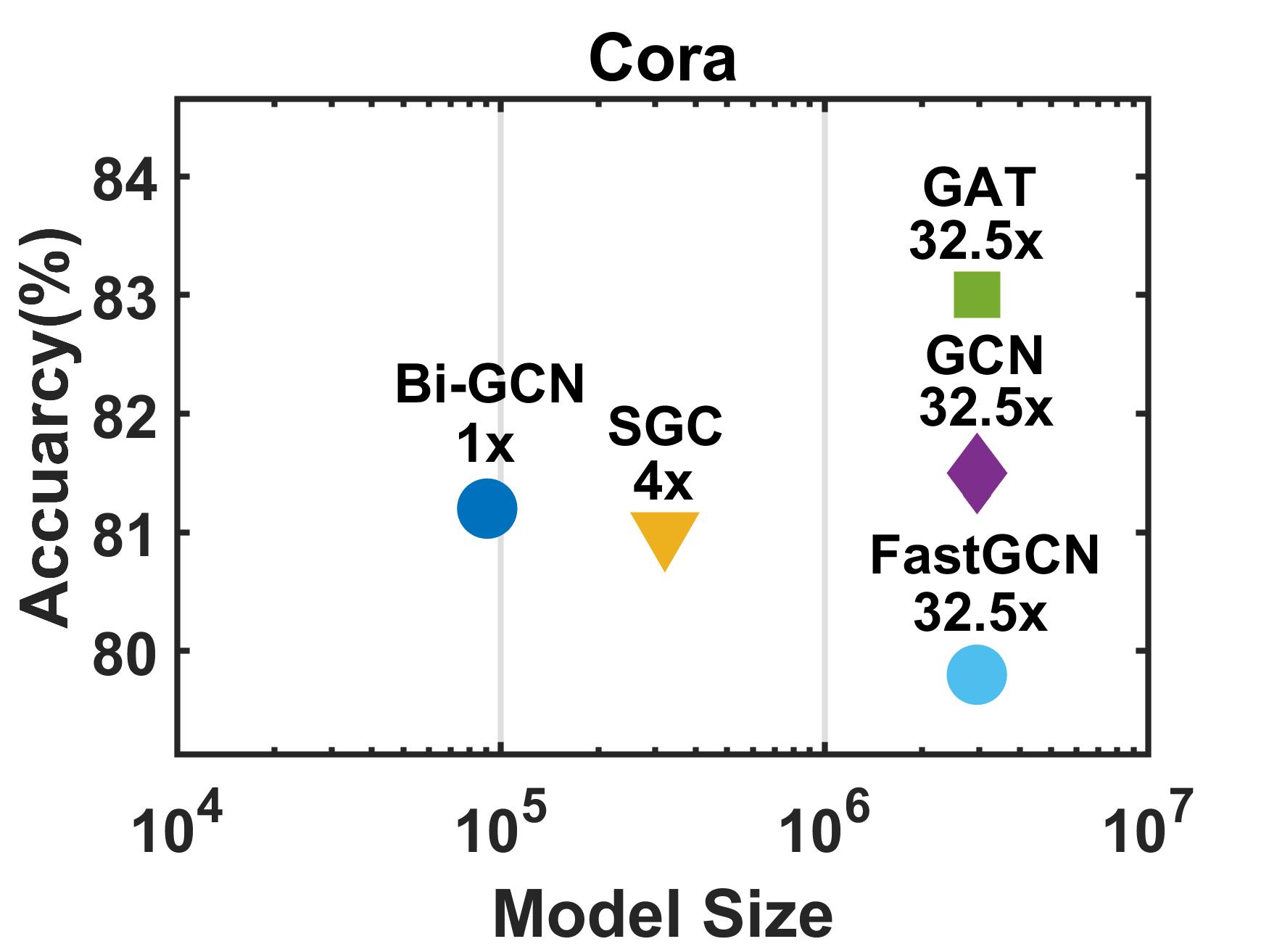}
  }
  \subfigure[] {
    \label{fig1:b}
    \includegraphics[width=0.45\columnwidth]{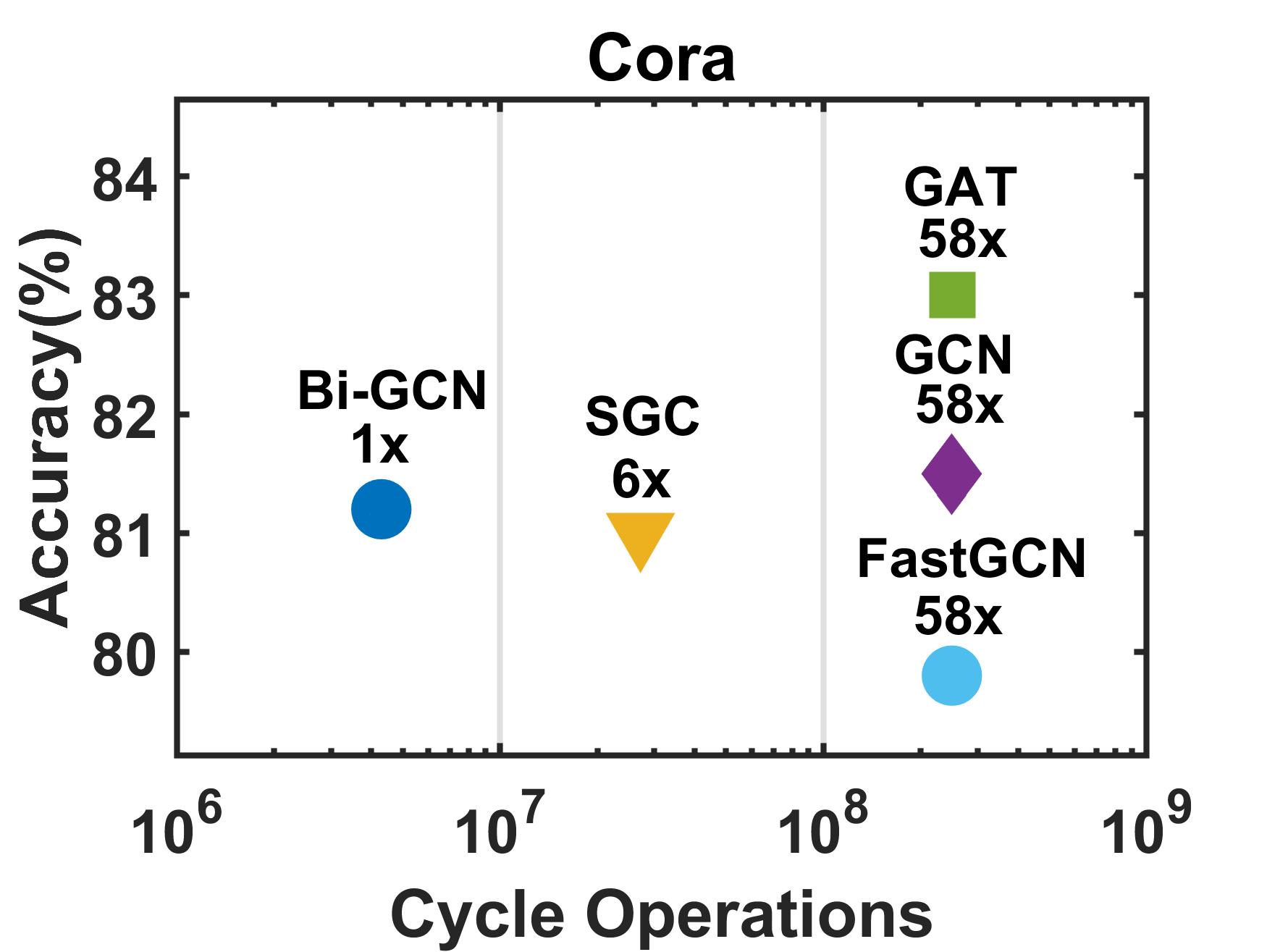}
  }
  \caption{Performances on the Cora dataset.
  Note that the model size is measured in bits.
  The number of cycle operations, which will be introduced in Sec. \ref{Sec-Analysis of Efficiency}, is employed to reflect the inference speed.
  Our Bi-GCN gives the fastest inference speed and the lowest memory consumption with comparable accuracy.
  }
  \label{fig1}
\end{figure}

Another feasible solution is compression, i.e., compressing the size of the input graph data and the GNN model to better utilize the limited memory and computational resources.
Certain approaches have been proposed to compress the convolutional neural networks (CNNs), such as designing shallow networks \cite{shallownetworks}, pruning \cite{compress2}, designing compact layers \cite{inception}, and quantizing the parameters \cite{Binarized-Neural-Networks}.
Among these approaches, quantization has been widely employed in practice, due to its excellent performance in reducing memory consumption and computational complexity.
Binarization \cite{Binarized-Neural-Networks,xnornet,bi-real}, a special type of quantization-based methods, has achieved great success in many CNN-based vision tasks, when a faster speed and lower memory consumption is desired.

Unfortunately, compared to the compression of CNNs, the compression of GNNs possesses unique challenges.
Firstly, since the input graph data is usually much larger than the GNN models, the compression of the loaded data demands more attention.
Secondly, the nodes tend to be similar to their neighbors in the high-level semantic space, while they tend to be different in the low-level feature space.
This characteristic is different from the grid-like data, such as images, videos, etc.
It requires the compressed GNNs to possess sufficient parameters for the representation learning.
At last, GNNs are generally shallow, e.g., the standard GCN \cite{gcn} only has two layers, which contain fewer redundancies.
Thus, the compression of GNNs is more challenging to be achieved.

To tackle the memory and complexity issues, SGC \cite{sgc}, an 1-layered GNN, compresses GCN \cite{gcn} by removing its nonlinearities and collapsing weight matrices between consecutive layers.
This 1-layered GNN can accelerate both the training and inference processes with comparable performance.
Although SGC compresses the network parameters, it does not compress the loaded data, which is the primary memory consumption when processing the graphs with GNNs.

In this paper, to alleviate the memory and complexity issue, we pioneer to propose a binarized GCN, named Binary Graph Convolutional Network (Bi-GCN), a simple yet efficient approximation of GCN \cite{gcn}, by binarizing the parameters and node attribute representations.
Specifically, the binarization of the weights is performed, by splitting them into multiple feature selectors and maintaining a scalar per selector to reduce the quantization errors further.
Similarly, the binarization of the node features can be carried out by splitting the node features and assigning an attention weight to each node.
With the employment of these additional scalars, more efficient information can be learned and retained efficiently.
After binarizing the weights and node features, the computational complexity and the memory consumptions, which are induced by the network parameters and input data, can be vastly reduced.
Since the existing binary back propagation method \cite{xnornet} has not considered the relationships among the binary weights, we also design a new back propagation method by tackling this issue.
An intuitive comparison between our Bi-GCN and the baseline methods is shown in Figure \ref{fig1}, which demonstrates that our Bi-GCN can achieve the fastest inference speed and lowest memory consumption with a comparable accuracy compared to the standard full-precision GNNs.

In general, our proposed Bi-GCN can reduce the redundancies in the node representations while maintaining the principle information.
When the number of layers increases, Bi-GCN also gives a more obvious reduction of the memory consumptions of the parameters and effectively alleviates the overfitting problem.
Besides, our binarization approach of Bi-GCN can be easily applied to other GNNs.
We introduce a general binarization approach to binarize other variants of GNN. 
To combine the binarization with other efficient techniques, like attention and sampling, we give the detailed binarized version of three of the most popular GNNs, i.e., GAT, GraphSAGE, and GraphSAINT, followed by the general binarization approach.
Experiments verify that these Bi-GNNs are also effective.

Similar to other compressed neural networks, our Bi-GCN and other Bi-GNNs may also meet the \textit{capacity problem}.
For example, we assume that a floating-point GCN achieves the peak performance for the semi-supervised classification task with exactly enough parameters, which indicates that the information storage capacity of this GCN is just appropriate for this task.
Under such circumstance, a direct binarization of this specific floating-point GCN may not function decently, because the storage capacity is decreased after the binarization.
To solve this problem, a practical solution is to increase the width of the hidden layer to empirically search for a suitable capacity.
However, it is still unclear why the searched width is appropriate to approximate the storage capacity of the floating-point GCN.

To tackle this \textit{capacity problem} theoretically, we propose an Entropy Cover Hypothesis to estimate the lower bound of the proper width of binary hidden layers of Bi-GNNs, compared to the floating-point GNNs.
According to the information theory, entropy and maximum discrete entropy can represent the actual amount of information and the storage capacity \cite{entropy_and_generalization_in_feedforward_nets}.
The Entropy Cover Hypothesis assumes that the storage capacity of the hidden layer in Bi-GNN should be no smaller than the amount of information of the hidden layer in a well-trained GNN.
Then, we can further conclude that the lower bound of the width of Bi-GNN hidden layers, which is equivalent to the storage capacity, should be no smaller than the maximum entropy of the hidden layers in the well-trained corresponding floating-point GNN.
The effectiveness of the proposed hypothesis can be proven experimentally.


The contributions are summarized as follows:
\begin{itemize}
\item We pioneer to propose a binarized GCN, named Binary Graph Convolutional Network (Bi-GCN), which can significantly reduce the memory consumptions by $\thicksim$31x for both the network parameters and input node attributes, and accelerate the inference by an average of $\thicksim$51x, on three citation networks, theoretically.
\item We design a new back propagation method to effectively train our Bi-GCN, by considering the relationships among the binary weights in the back propagation process.
\item We introduce a general method to generalize our binarization approach to other GNNs. With respect to the significant memory reductions and accelerations, our binarized GNNs can also give comparable performance to the floating-point GNNs on seven node classification tasks.
\item We propose an Entropy Cover Hypothesis, which theoretically analyzes the \textit{capacity problem}, to describe the lower bound of the proper width for binary hidden layers of Bi-GNNs.
\end{itemize}

A preliminary version of this paper was published in \cite{bi-gcn}. 
This paper significantly improves \cite{bi-gcn} in the following aspects. 
(i) We introduce a general binarization approach for other variants of GNNs to combine the advantages of binarization with other efficient techniques, while \cite{bi-gcn} mainly focuses on the design of Bi-GCN.
(ii) The Entropy Cover Hypothesis, which gives a theoretical explanation for the \textit{capacity problem}, is proposed to describe the lower bound of the proper width of binary hidden layers of Bi-GNN according to a well-trained floating-point GNN. 
(iii) More evaluations are conducted with the newly proposed benchmark, OGB \cite{ogb}, which further demonstrate the effectiveness of the proposed Bi-GCN and Bi-GNNs.

The rest of this paper is organized as follows. 
Section 2 reviews the related literatures of the sampling techniques for GNNs, as well as the binarization methods for CNNs.
Section 3 defines the mathematical notations and introduces the popular Graph Convolution Network \cite{gcn}.
Section 4 presents the proposed Bi-GCN and the generalized binarization approach.
Section 5 analyzes the efficiencies of the Bi-GNNs obtained via our binarization approach.
Section 6 introduces the proposed Entropy Cover Hypothesis.
Section 7 discusses the experimental results on seven node classification datasets.
At last, section 8 concludes this paper.

\section{Related Work}
\subsection{Sampling Techniques for GNNs}

Traditional GNNs \cite{gcn1, gcn2, gcn} usually aim to process the relatively small datasets, thus they usually load the entire graph in the training and testing process.
Apparently, the memory cannot be infinitely provided, when the scale of the graph grows larger.
Sampling \cite{graphsage, pinsage, stgcn, fastgcn, LADIES, zeng2019accurate, cluster-gcn, graphsaint} is an effective mechanism which allows GNNs to process larger graphs with limited memory.

As the first attempt, the neighbor sampling method in GraphSAGE \cite{graphsage} samples a local neighborhood for each node and then aggregates the neighborhood to obtain its local features.
PinSAGE \cite{pinsage} further improves the neighbor sampling of \cite{graphsage} by considering the importance of different neighbors, which is determined by the visit count of random walk.
However, the total number of the sampled neighbors will grow exponentially when the depth of the GNN model grows, which is named as the \textit{neighbor explosion} problem.
To alleviate this issue, VRGCN \cite{stgcn} reduces the neighborhood size of \cite{graphsage} by selecting only two neighbors from the previous layer.
The historical activations of each node are utilized to maintain the stability of the estimator.

Different from the neighbor sampling mechanism, FastGCN \cite{fastgcn} directly samples a subgraph via an importance sampling strategy in each layer to accelerate the training process.
Since this layer-wise graph sampling method does not obey the scheme of neighbor sampling, it can avoid \textit{neighbor explosion}.
However, the sampled nodes in FastGCN \cite{fastgcn} are not necessarily connected.
Therefore, the sampled subgraph may be too sparse to represent the local neighborhood and this strategy tends to sacrifice the classification accuracy.
LADIES \cite{LADIES} further improves this strategy in FastGCN \cite{fastgcn} by generating samples from the neighborhood of the node which is already sampled in the previous layer.

Instead of sampling the layer-wise subgraphs, graph sampling methods generate a set of subgraphs as mini-batches and train GNN directly on the sampled subgraphs.
Frontier sampling method \cite{zeng2019accurate} generates subgraphs via a multi-dimensional random walk algorithm to ensure the connectivity in each subgraph.
GraphSAINT \cite{graphsaint} proposes an edge sampling method with low variance and a graph sampling training framework with unbiased estimators.
However, in these graph sampling methods, local connections tend to be more sparse when the scale of the graph grows.
On the contrary, graph clustering algorithms is utilized in ClusterGCN \cite{cluster-gcn} to partition the graph into several subgraphs.
To better utilize the between-subgraph links, subgraphs are randomly selected to form mini-batches.
Unfortunately, this partition algorithm demands additional computations for the clustering.

Note that the idea of sampling is also utilized to alleviate \textit{over-smoothing}.
The \textit{over-smoothing} phenomenon is firstly introduced in \cite{deeperinsights}, i.e., the features of the nodes in a connected component will become indistinguishable when the depth of GCN increases.
DropEdge \cite{dropedge} randomly drops out/removes edges at a specific rate in the training process.
It makes the node connections more sparse and hence alleviates the potential over-smoothing.

\subsection{Binarization Methods for CNNs}

Despite achieving great successes in many fields, CNNs also suffer from specific issues such as over-parametrization and high computational costs.
Binarization, which attempts to binarize the weights and/or the activations in the neural networks, is a promising network compression technique to reduce the consumptions of memory and computations.
BinaryConnect \cite{BinaryConnect} binarizes the network parameters and replaces most of the floating-point multiplications in the inference process with floating-point additions.
BinaryNet \cite{Binarynet} further binarizes the activation function and employes the XNOR (not-exclusive-OR) operations instead of the floating-point additions.
XNOR-Net \cite{xnornet} proposes a binarization method with scalars and successfully applies it to the famous CNNs, such as Residual Networks \cite{resnet} and GoogLeNet \cite{googlenet}.

\section{Preliminaries}
\subsection{Notations}
Here, we define the notations utilized in this paper.
We denote an undirected attributed graph as $\mathcal{G}=\left\{\mathcal{V},\mathcal{E},\mathcal{X}\right\}$ with a set of vertices, $\mathcal{V} =\{v_i\}_{i=1}^N$, and a set $\mathcal{E}=\{e_i\}_{i=1}^E$ of edges,. 
Each node $v_i$ contains a feature $X_i\in\mathbb{R}^d$.
$X\in\mathbb{R}^{N\times d}$ is the collection of all the features in all the nodes. 
$A=[a_{ij}] \in\mathbb{R}^{N\times N}$ represents the adjacency matrix, which reveals the relationships between each pair of vertices, i.e., the topology information of $\mathcal{G}$.
$d_i=\sum_j a_{ij}$ stands for the degree of node $v_i$, and $D=diag(d_1, d_2, \dots, d_n)$ represents the degree matrix corresponding to the adjacency matrix $A$. 
Then, $\hat{A} = A + I$ is the adjacency matrix of the original topology with self-loops, and $\hat{D}$ is its corresponding degree matrix with $\hat{D}_{ii} = \sum_j\hat{a}_{ij}$.
Note that we employ the superscript ``$(l)$'' to represent the $l$-th layer, e.g., $H^{(l)}$ is the input node features to the $l$-th layer.

\subsection{Graph Convolutional Network}
Graph Convolutional Network (GCN) \cite{gcn} has become the most popular graph neural network in the past few years.
Since our binarization approach takes GCN as the basis GNN, we briefly review GCN here.

Given an undirected graph $\mathcal{G}$, the graph convolution operation can be described as
\begin{equation}
  H^{(l+1)} = \sigma(\tilde{A}H^{(l)}W^{(l)}),
  \label{gcn-propagation}
\end{equation}
where $\tilde{A}=\hat{D}^{-\frac{1}{2}}\hat{A}\hat{D}^{-\frac{1}{2}}$ is a sparse matrix, and $W^{(l)}\in\mathbb{R}^{d_{in}^{(l)}\times d_{out}^{(l)}}$ contains the learnable parameters.
Note that $H^{(l+1)}$ is the output of the $l$-th layer as well as the input of the $(l+1)$-th layer, and $H^{(0)} = X$.
$\sigma$ is the non-linear activation function, e.g., ReLU.

From the perspective of spatial methods, the graph convolution layer in GCN can be decomposed into two steps, where $\tilde{A}H^{(l)}$ is the aggregation step and $H^{(l)}W^{(l)}$ is the feature extraction step.
The aggregation step tends to constrain the node attributes in the local neighborhood to be similar.
After the aggregation, the feature extraction step can easily extract the commonalities between the neighboring nodes.

GCN typically utilizes a task-dependent loss function, e.g., the cross-entropy loss for the node classification tasks, which is defined as
\begin{equation}
  \mathcal{L} = -\sum_{v_i\in\mathcal{V}^{label}}\sum_{c=1}^{C}Y_{i,c}log(\tilde{Y}_{i,c}),
  \label{gcn-loss}
\end{equation}
where $\mathcal{V}^{label}$ stands for the set of the labeled nodes, $C$ denotes the number of classes, $Y$ represents the ground truth labels, and $\tilde{Y}=softmax(H^{(L)})$ are the predictions of the $L$-layered GCN.

\section{Binary Graph Convolutional Network}
\label{Sec-BiGCN}
This section presents the proposed Binary Graph Convolution Network (Bi-GCN), a binarized version of the standard GCN \cite{gcn}. 
As mentioned in the previous section, a graph convolution layer can be decomposed into two steps, aggregation and feature extraction.
In Bi-GCN, we only focus on binarizing the feature extraction step, because the aggregation step possesses no learnable parameters (which yields negligible memory consumption) and it only requires a few calculations (which can be neglected compared to the feature extraction step).
Therefore, the aggregation step of the original GCN is maintained.
For the feature extraction step, we binarize both the network parameters and node features to reduce the memory consumptions.
To reduce the computational complexities and accelerate the inference process, the XNOR (not-exclusive-OR) and bit count operations are utilized, instead of the traditional floating-point multiplications.
Then, we design an effective back-propagation algorithm for training our binarized graph convolution layer.
At last, we introduce a general method to apply our binarization approach to other GNNs and give examples of three commonly used GNNs, i.e., GAT, GraphSAGE, and GraphSAINT.

\begin{figure*}[t]
  \centering
  \includegraphics[width=1.8\columnwidth]{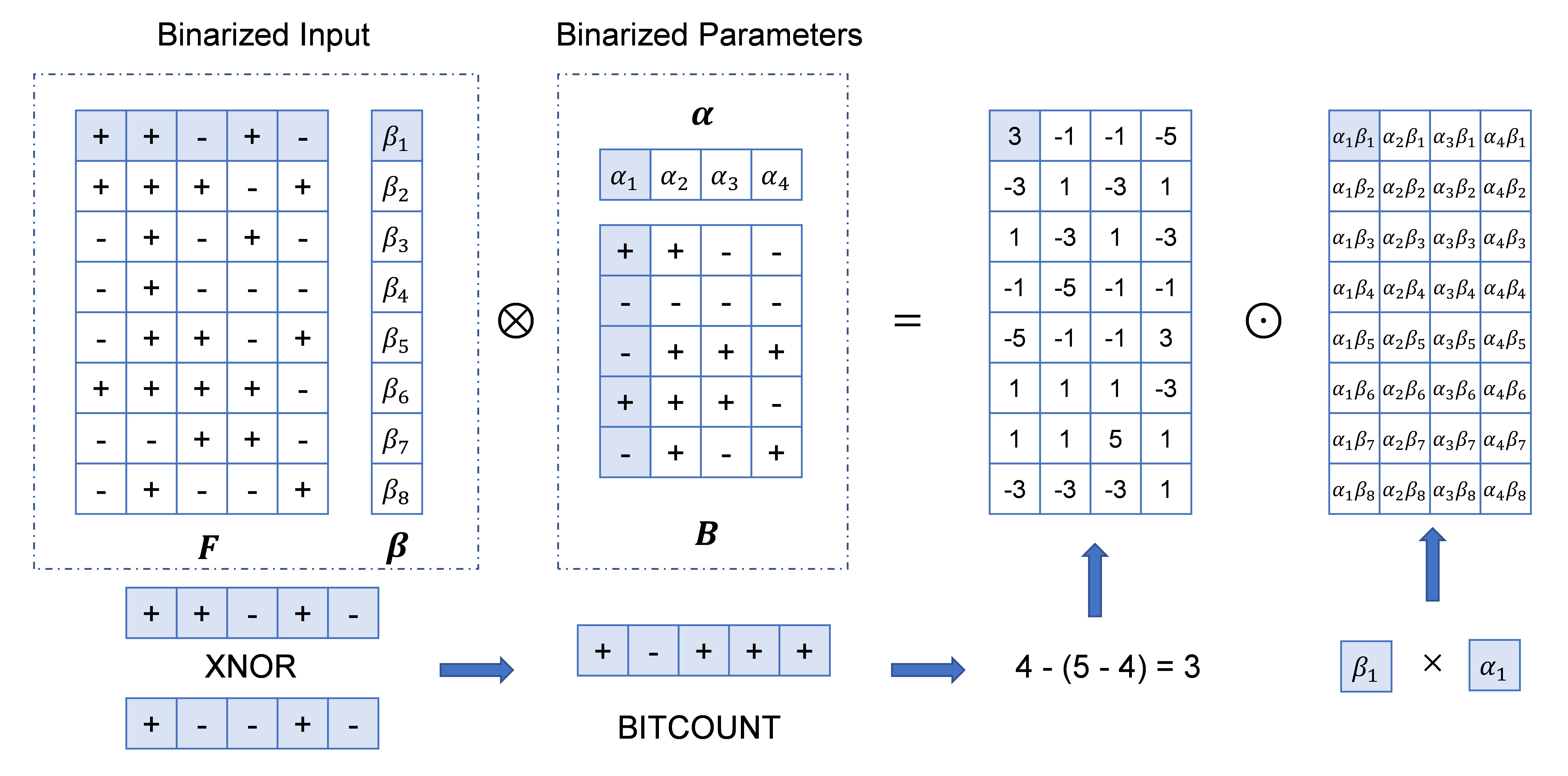}
  \caption{An example of binary feature extraction step.
  Both the input features and parameters will be binarized to binary matrices.
  $\otimes$ denotes the binary matrix multiplication defined in Sec. \ref{Sec-BiGCN} and $\odot$ represents the element-wise multiplication.
  }
  \label{fig2}
\end{figure*}

\subsection{Vector Binarization}

  Firstly, we introduce the vector binarization approach, which is a vital part of our binarization process.
  Considering that there exists a vector $V=(V_1, V_2,...,V_t)$, we aim to obtain its binarized approximation with a binary vector $V_B = \{-1, 1\}^t$ and a real-valued scalar $\alpha$, such that $V\approx\alpha{V_B}$.
  This approximation can be formulated as
  \begin{equation}
    J_v(V_B, \alpha) = ||V-\alpha V_B||^2_2.
    \label{dotproduct-cost1}
  \end{equation}
  By minimizing the above optimization problem, the optimal solution \cite{xnornet} can be computed via
  \begin{equation}
    V_B^*=sign(V),
  \end{equation}
  \begin{equation}
    \alpha^*=\frac{1}{t}||V||_1,
  \end{equation}
  where $sign(\cdot)$ is the signum function which extracts the sign of a real number.
  
  Then, if there exists another vector $I=(I_1, I_2,...,I_t)$, the inner product of $I$ and $V$ can be approximated via
  \begin{equation}
    I \cdot V \approx \alpha^* I \cdot V_B^*,
    \label{dotproduct-VB}
  \end{equation}
  where $\cdot$ denotes the vector inner product.
  If a further binarization to the vector $I$ is desired to compress this inner product, it can be achieved via $I \cdot V \approx \alpha\beta I_B \cdot V_B$, where $I_B$ is a binary vector and $\beta$ is a scalar.
  If we intend to minimize the straightforward approximation error $|I \cdot V - \alpha\beta I_B \cdot V_B|$ and compute the optimal solution to this optimization problem, an optimal solution, $|I_B \cdot V_B|=1$ and $\alpha\beta = sign(I_B \cdot V_B) I \cdot V$, can easily be calculated.
  Unfortunately, this solution possesses a strong dependency on the value of $I \cdot V$, and it tends to lose a large amount of information of the original vectors.
  To alleviate this issue, we define the approximation problem of the inner product $I\cdot V$ as
  \begin{equation}
    J_{ip}(\alpha,\beta, I_B, V_B) = ||I\odot V-\alpha\beta I_B \odot V_B||^2_2,
    \label{dotproduct-cost2}
  \end{equation}
  where $\odot$ denotes the element-wise product. 
  Similar to Eq. \ref{dotproduct-cost1}, the optimal solution can be calculated via
  \begin{equation}
    \alpha^*=\frac{1}{t}||V||_1,
  \end{equation} 
  \begin{equation}
    \beta^*=\frac{1}{t}||I||_1,
  \end{equation}
  \begin{equation}
    V_B^*=sign(V),
  \end{equation} 
  \begin{equation}
    I_B^*=sign(I).
  \end{equation} 
  Then, Eq. \ref{dotproduct-VB} can be reformed to
  \begin{equation}
    I \cdot V \approx \alpha^*\beta^* I_B^* \cdot V_B^*.
    \label{dotproduct-VBIB}
  \end{equation}
  Eq. \ref{dotproduct-VBIB} is essentially the result of binarizing both $I$ and $V$ according to our vector binarizing algorithm.

\subsection{Binarization of the Feature Extraction Step}

Based on the vector binarization algorithm, we can perform the binarization to the feature extraction step $Z^{(l)}=H^{(l)}W^{(l)}$ in the graph convolution shown in Eq. \ref{gcn-propagation}.
Note that for this feature extraction (matrix multiplication) step, we adopt the bucketing \cite{qsgd} method to generalize the binary inner product operation to the binary matrix multiplication operation.
Specifically, we split the matrix into multiple buckets of consecutive values with a fixed size and perform the scaling operation separately.

\subsubsection{Binarization of the Parameters}
Since each column of the parameter matrix of the $l$-th layer $W^{(l)}$ serves as a feature selector in the computation of $Z^{(l)}$, each column of $W^{(l)}$ is splitted as a bucket.
Let $\alpha^{(l)}=(\alpha_1^{(l)}, \alpha_2^{(l)}, ..., \alpha_{d_{out}^{(l)}}^{(l)})$, which are the scalars for each bucket.
Then, the binarization of $W^{(l)}$ can be achieved based on the buckets and their corresponding scalars.
Note that the value of scalar $\alpha^{(l)}$ actually determines the importance of each feature and can thus be considered as a feature attention.

Let $B^{(l)}=(B_1^{(l)},B_2^{(l)},...,B_{d_{out}^{(l)}}^{(l)})\in\{-1, 1\}^{d_{in}^{(l)}\times d_{out}^{(l)}}$ be the binarized buckets of $W^{(l)}$.
Then, based on the vector binarization algorithm, the optimal $B^{(l)}$ and $\alpha^{(l)}$ can be easily calculated by
\begin{equation}
  B_j^{(l)}=sign(W^{(l)}_{:,j}),
  \label{B}
\end{equation}
\begin{equation}
  \alpha_j^{(l)}=\frac{1}{d_{out}^{(l)}}||W^{(l)}_{:,j}||_1,
  \label{alpha}
\end{equation}
where $W^{(l)}_{:,j}$ represents the $j$-th column of $W^{(l)}$. 
It can be approximated via
\begin{equation}
  W^{(l)}_{:,j}\approx\tilde{W}^{(l)}_{:,j}=\alpha_j^{(l)} B^{(l)}_j.
  \label{app-W}
\end{equation}
Based on Eq. \ref{app-W}, the graph convolution operation with binarized weights can then be described as
\begin{equation}
  H^{(l+1)} \approx H^{(l+1)}_{p} = \sigma(\tilde{A}H^{(l)}\tilde{W}^{(l)}),
\end{equation}
where $H^{(l+1)}_{p}$ is the binary approximation of $H^{(l+1)}$ with the binarized parameters $\tilde{W}^{(l)}$.
The binarization of the parameters can reduce the memory consumption by a factor of $\thicksim$30x, compared to the parameters with full precision.

\subsubsection{Binarization of the Node Features}
Due to the over-smoothing issue \cite{deeperinsights} induced by the current graph convolution operation, current GNNs are usually shallow, e.g., the vanilla GCN only contains 2 graph convolution layers.
Although the future GNNs may possess a larger model, the data sizes of commonly employed attributed graphs are usually much larger than the current model size.
To reduce the memory consumption of the input data, which is mostly induced by the node features, we also perform binarization to the node features which will be processed by the graph convolutional layers.

To binarize the node features, we split $H^{(l)}$ into row buckets based on the constraints of the matrix multiplication to compute $Z^{(l)}$, i.e., each row of $H^{(l)}$ will conduct an inner product with each column of $W^{(l)}$.
Let $\beta^{(l)}=(\beta_1^{(l)}, \beta_2^{(l)}, ..., \beta_{N}^{(l)})$ denote the scalars for each bucket in $H^{(l)}$.
Let $F^{(l)}=(F_1^{(l)}; F_2^{(l)}; ...; F_N^{(l)})\in\{-1, 1\}^{N\times d_{in}^{(l)}}$ be the binarized buckets.
Then, with the vector binarization algorithm, the optimal $\beta$ and $F$ can be computed by

\begin{equation}
  \beta_i^{(l)}=\frac{1}{N}||H_{i,:}^{(l)}||_1,
  \label{beta}
\end{equation}
\begin{equation}
  F_i^{(l)}=sign(H^{(l)}_{i,:}),
  \label{F}
\end{equation}
where $H_{i,:}^{(l)}$ represents the $i$-th row of $H^{(l)}$. 
Then, the binary approximation of $H^{(l)}$ can be obtained via
\begin{equation}
  H^{(l)}_{i,:}\approx\tilde{H}^{(l)}_{i,:}=\beta_i^{(l)} F^{(l)}_i.
\end{equation}
Intuitively, $\beta$ can be considered as the node-weights for the feature representations.
At last, the graph convolution operation with binarized weights and node features can be formulated as
\begin{equation}
  H^{(l+1)} \approx H^{(l+1)}_{ip} = \tilde{A}\tilde{H}^{(l)}\tilde{W}^{(l)}.
  \label{H-ip}
\end{equation}

Note that this binarization of the node features, i.e., the input of the graph convolutional layer, also possesses the ability of activation.
Thus we do not employ specific activation functions (such as ReLU).
Similar to the binarization of the weights, the memory consumption of the loaded attributed graph data can be reduced by a factor of $\thicksim$30x compared to the vanilla GCN.

\subsubsection{Binary Operations} 
With the binarized graph convolutional layers, we can accelerate the calculations by employing the XNOR and bit-count operations instead of the floating-point additions and multiplications. 
Let $\zeta^{(l)}$ represent the approximation of $Z^{(l)}$. Then,
\begin{equation}
  Z_{ij}^{(l)}\approx\zeta_{ij}^{(l)} =\beta_i^{(l)}\alpha_j^{(l)} F_{i,:}^{(l)}\cdot B_{:,j}^{(l)}.
  \label{cdot-op}
\end{equation}
Since each element of $F^{(l)}$ and $B^{(l)}$ is either -1 or 1, the inner product between these two binary vectors can be replaced by the binary operations, i.e., XNOR and bit count operations.
Then, Eq. \ref{cdot-op} can be re-written as
\begin{equation}
  \zeta_{ij}^{(l)} =\beta_i^{(l)}\alpha_j^{(l)} F_{i,:}^{(l)}\circledast B_{:,j}^{(l)},
  \label{bi-op}
\end{equation}
where $\circledast$ denotes a binary multiplication operation using the XNOR and a bit count operations.
The detailed process is illustrated in Figure \ref{fig2}.
Therefore, the graph convolution operation in the vanilla GCN can be approximated by
\begin{equation}
  H^{(l+1)} \approx H^{(l+1)}_{b} = \tilde{A}\zeta^{(l)},
\end{equation}
where $\zeta^{(l)}$ is calculated via Eq. \ref{bi-op} and $H^{(l+1)}_{b}$ is the final output of the $l$-th layer with the binarized parameters and inputs.
By employing this binary multiplication operation, the original floating-point calculations can be replaced with identical number of  binary operations and a few extra floating-point calculations.
It will significantly accelerate the processing speed of the graph convolutional layers.

\begin{algorithm}[t]
  \caption{Back propagation process for training a binarized graph convolutional layer}
  \label{alg1}
  \renewcommand{\algorithmicrequire}{\textbf{Input:}}
  \renewcommand{\algorithmicensure}{\textbf{Output:}}
  \begin{algorithmic}[1]
  \REQUIRE Gradient of the layer above $\frac{\partial{\mathcal{L}}}{\partial{H^{(l+1)}}}$
  \ENSURE Gradient of the current layer $\frac{\partial{\mathcal{L}}}{\partial{H^{(l)}}}$
  \STATE Calculate the gradients of $\tilde{W}^{(l)}$ and $\tilde{H}^{(l)}$
  
  {$\frac{\partial{\mathcal{L}}}{\partial{\zeta^{(l)}}}=\tilde{A}^T\cdot\frac{\partial{\mathcal{L}}}{\partial{H^{(l+1)}}}$}

  {$\frac{\partial{\mathcal{L}}}{\partial{\tilde{W}^{(l)}}}=(\tilde{H}^{(l)})^T\cdot\frac{\partial{\mathcal{L}}}{\partial{\zeta^{(l)}}}$}
  
  {$\frac{\partial{\mathcal{L}}}{\partial{\tilde{H}^{(l)}}} = \frac{\partial{\mathcal{L}}}{\partial{\zeta^{(l)}}}\cdot\tilde{W}^{(l)}$}
  
  \STATE Calculate $\frac{\partial{\mathcal{L}}}{\partial{H^{(l)}}}$ via Eq. \ref{bp-H}
  \STATE Calculate $\frac{\partial{\mathcal{L}}}{\partial{W^{(l)}}}$ via Eq. \ref{bp-W}
  \STATE Update $\tilde{W}^{(l)}$ with the gradient $\frac{\partial{\mathcal{L}}}{\partial{W^{(l)}}}$
  \STATE \textbf{return} $\frac{\partial{\mathcal{L}}}{\partial{H^{(l)}}}$
  \end{algorithmic}
\end{algorithm}

\subsection{Binary Gradient Approximation Based Back Propagation}
The critical parts of our training process include the choice of the loss function and the back-propagation method for training the binarized graph convolutional layer.
The loss function employed in our Bi-GCN is the same as the vanilla GCN, as shown in Eq. \ref{gcn-loss}.
Since the existing back-propagation method \cite{xnornet} has not considered the relationships among the binary weights, to perform back-propagation for the binarized graph convolutional layer, the gradient calculation is desired to be newly designed.

To calculate the actual propagated gradient for the $l$-th layer, the binary approximated gradient $\frac{\partial{\mathcal{L}}}{\partial{\tilde{H}^{(l)}}}$ is employed to approximate the gradient of the original one as \cite{Binarized-Neural-Networks,xnornet},
\begin{equation}
  \frac{\partial{\mathcal{L}}}{\partial{H^{(l)}}} \approx\frac{\partial{\mathcal{L}}}{\partial{\tilde{H}^{(l)}}}\mathbbm{1}_{|\frac{\partial{\mathcal{L}}}{\partial{\tilde{H}^{(l)}}}|<1}.
  \label{bp-H}
\end{equation}
Note that $\mathbbm{1}_{|r|<1}$ is the indicator function, whose value is 1 when $|r|<1$, and vice versa.
This indicator function serves as a hard ${\rm tanh}$ function which preserves the gradient information.
If the absolute value of the gradients becomes too large, the performance will be degraded.
Thus, the indicator function also serves to kill certain gradients whose absolute value becomes too large.

The gradient of network parameters is computed via another gradient calculation approach.
Here, a full-precision gradient is employed to preserve more gradient information.
If the gradient of the binarized weights $\frac{\partial{\mathcal{L}}}{\partial{\tilde{W}^{(l)}}}$ is obtained, $\frac{\partial{\mathcal{L}}}{\partial{W_{ij}^{(l)}}}$ can then be calculated as
\begin{equation}
  \begin{aligned}
  \frac{\partial{\mathcal{L}}}{\partial{W_{ij}^{(l)}}} =& \frac{\partial{\mathcal{L}}}{\partial{\tilde{W}_{:,j}}^{(l)}}\cdot\frac{\partial{\tilde{W}_{:,j}^{(l)}}}{\partial{W_{ij}^{(l)}}}\\
  =&\frac{1}{d_{in}^{(l)}}{B_{ij}^{(l)}}\sum_k\frac{\partial{\mathcal{L}}}{\partial{\tilde{W}_{kj}^{(l)}}}\cdot{B_{kj}^{(l)}}+\alpha_j^{(l)}\cdot\frac{\partial{\mathcal{L}}}{\partial{\tilde{W}_{ij}^{(l)}}}\cdot\frac{\partial{B_{ij}^{(l)}}}{\partial{W_{ij}^{(l)}}}.
  \end{aligned}
  \label{bp-W}
\end{equation}

To compute the gradient for the sign function $sign(\cdot)$, the straight-through estimator (STE) function \cite{st-estimating} is employed, where $\frac{\partial{sign(r)}}{\partial{r}} = \mathbbm{1}_{|r|<1}$. 
The back-propagation process is summarized in Algorithm \ref{alg1}.

\subsection{Generalization to other Bi-GNNs}
Here, we introduce the generalization of our binarization approach to other popular GNN variants.
The general method to binarize a GNN layer consists of three steps.
Firstly, a standard batch normalization \cite{bn} (with zero mean and variance being one) is utilized for the input of the layer to keep the balance of -1 and 1 after binarization.
Then, the limited storage capacity can be utilized as much as possible.
Secondly, the input $H^{(l)}$ and the parameters $\mathbf{\Theta}^{(l)}$ of the layer are binarized, where the binary operations are utilized instead of matrix multiplications.
Thirdly, the original non-linear function, e.g., ReLU, is removed, because of two reasons.
1) The sign function in our binarization method already serves as a non-linear activation.
2) The family of ReLU-like activation functions tends to preserve the positive values and largely suppress the negative values, which obviously changes the distribution of positive and negative values and conflicts with the adopted sign function. 

Specifically, we will present three detailed examples, i.e., binarize three of the most famous GNN variants, including GAT, GraphSAGE and GraphSAINT.
By binarizing these fundamental yet practical techniques, such as attention and sampling in GNNs, different Bi-GNNs can be constructed for different practical scenarios. 

\subsubsection{Bi-GAT}

Graph Attention Network \cite{gat}, which is a popular variant of GNNs, learns a weighted aggregation function by applying the self-attention strategy to node features.
A typical GAT's convolutional layer is defined as
\begin{equation}
  h^{(l+1)}_{i} = \sigma(\sum_{j\in\mathcal{N}_i} { \alpha_{ij} W h^{(l)}_{j}}),
  \label{gat-conv}
\end{equation}
where $\alpha_{ij}$ is the attention score calculated via
\begin{equation}
  \alpha_{ij} = {softmax}\left({LeakyReLU} \left(a^T (Wh_i||Wh_j)\right)\right).
  \label{gat-att}
\end{equation}
Then, by utilizing our binarization approach, the Bi-GAT's convolutional layer can be obtained as
\begin{equation}
  h^{(l+1)}_{i} \approx \sum_{j\in\mathcal{N}_i} { \alpha_{ij} \cdot \tilde{W} \circledast \tilde{h}^{(l)}_{j}},
  \label{bigat-conv}
\end{equation}
where $\alpha_{ij}$ is calculated by replacing $Wh$ in Eq. \ref{gat-att} with its binary version $\tilde{W} \circledast \tilde{h}$.
Note that we do not binarize the attention parameters $a^T$, since it is also efficient, i.e., the calculations and memory consumptions introduced by $\alpha$ are approximated to those of our binarization version of $W$ and $h$.


\subsubsection{Bi-GraphSAGE}

GraphSAGE \cite{graphsage} proposes an inductive learning scheme, i.e., neighbor sampling and aggregation, for the GNN models.
Four kinds of aggregators are utilized, i.e., mean, LSTM, pooling, and the aggregator in GCN, in its aggregation process.
All of them tend to give similar performances in the evaluations.
In the subsequent literatures, the mean aggregator is commonly utilized in GraphSAGE, which is constructed as

\begin{equation}
  h^{(l+1)}_{i} = {ReLU} \left(W_{\theta}^{(l)} \cdot h^{(l)}_{i} + \frac{1}{|\mathcal{N}_i|}\sum_{j\in\mathcal{N}_i} W_{n}^{(l)} \cdot h^{(l)}_{j}\right), 
  \label{sage-conv}
\end{equation}
where $\mathcal{N}_i$ represents the set of neighbors of node $v_i$, and $|\mathcal{N}_i|$ denotes the total number of the neighbors.
At last, for an $L$-layered GraphSAGE, $h_{i}^{(L)}$ is utilized to generate the prediction of node $v_i$.
According to our binarization approach, the Bi-GraphSAGE's convolutional layer is formed as
\begin{equation}
  h^{(l+1)}_{i} \approx \tilde{W}_{\theta}^{(l)} \circledast \tilde{h}^{(l)}_{i} + \frac{1}{|\mathcal{N}_i|}\sum_{j\in\mathcal{N}_i} \tilde{W}_{n}^{(l)} \circledast \tilde{h}^{(l)}_{j},
  \label{bisage-conv}
\end{equation}
where $\tilde{h}^{l}$ is the binary node representations, $\tilde{W}_{\theta}$, and $\tilde{W}_{n}$ are the parameters to be learned, and $\circledast$ represents the binary operations in Sec 4.2. 
Similar to our Bi-GCN, we do not preserve the original non-linear function.

\subsubsection{Bi-GraphSAINT}
GraphSAINT \cite{graphsaint} proposes a graph sampling-based GNN training framework.
It builds mini-batches with a set of subgraphs and constructs a full GNN model on each mini-batch.
This graph sampling framework can be effectively applied to a variety of GNN models, e.g., GraphSAGE \cite{graphsage}, GAT \cite{gat}, JK-Net \cite{jk-net}, etc.
Since the binarization versions of GraphSAGE and GAT are already introduced, JK-Net is employed as the GNN model for GraphSAINT as our baselines, for the diversity of model selection.

Jumping knowledge networks (JK-Net) \cite{jk-net} proposes a layer aggregation operation to mix the representations of different hops in the local neighborhood.
For an $L$-layered JK-Net,
\begin{equation}
  Z_{i} = LA\left(h^{(1)}_i, h^{(2)}_i,...,h^{(L)}_i\right),
  \label{jknet-conv}
\end{equation}
which is utilized to help the prediction of node $v_i$.
Note that $h^{(l)}_i$ is the representation of the $l$-th GNN layer and $LA(\cdot)$ is the layer aggregation function, which can be implemented as concatenation, pooling, LSTM, etc. 
Here, the commonly used concatenation operation is utilized, i.e.,
\begin{equation}
  Z_{i} = W_{LA}\left(h^{(1)}_i || h^{(2)}_i || ... || h^{(L)}_i\right),
  \label{jknet-concat}
\end{equation}
where $||$ represents the concatenation operation.
Similar to the feature extraction step, the node features and weight parameters are binarized via
\begin{equation}
  Z_{i} = \tilde{W}_{LA}\circledast\left(\tilde{h}^{(1)}_i || \tilde{h}^{(2)}_i || ... || \tilde{h}^{(L)}_i\right).
  \label{bijknet-concat}
\end{equation}
In the employed GraphSAINT, the above version of JK-Net is utilized where $h_i^{(l)}$ is calculated by the GraphSAGE layer introduced in Eq. \ref{sage-conv}.

\section{Analysis of Efficiency}
\label{Sec-Analysis of Efficiency}

In this section, we take our Bi-GCN as an example to provide a theoretical analysis of the efficiency of our Bi-GNNs, i.e., the compression ratio of the model size and the loaded data size, as well as the acceleration ratio, compared to the full-precision (32-bit floating-point) GCN.
The compression and acceleration ratios of other Bi-GNNs can be analyzed in a similar manner.

\subsection{Model Size Compression}
Let the parameters of each layer in the full-precision GCN be denoted as $W^{(l)}\in\mathbb{R}^{d_{in}^{(l)}\times d_{out}^{(l)}}$, which contains $(d_{in}^{(l)}\times d_{out}^{(l)})$ floating-point parameters.
On the contrary, the $l$-th layer in our Bi-GCN only contains $(d_{in}^{(l)}\times d_{out}^{(l)})$ binary parameters and $d_{out}^{(l)}$ floating-point parameters.
Therefore, the size of the parameters can be reduced by a factor of 
\begin{equation}
  PC^{(l)} =\frac{32d_{in}^{(l)}d_{out}^{(l)}}{d_{in}^{(l)}d_{out}^{(l)}+32d_{out}^{(l)}} = \frac{32d_{in}^{(l)}}{d_{in}^{(l)}+32}.  
  \label{pc-l}
\end{equation}

According to Eq. \ref{pc-l}, the compression ratio of the parameters for the $l$-th layer depends on the dimension of the input node features. 
For example, a 2-layered Bi-GCN, whose hidden layer contains 64 neurons, can achieve a $\thicksim$31x model size compression ratio compared to the full-precision GCN on the Cora dataset. 
Although the memory consumption of the network parameters is smaller than the input data for the vanilla GCN, our binarization approach still contributes.
Currently, many efforts have already been made to construct deeper GNNs \cite{deepgcn,dropedge,fastanddeepgcn}.
As the number of layers increases, the reductions on the memory consumptions will become much larger, and this contribution will become more significant.

\begin{table}[h]
  \centering
  \caption{Memory Consumptions and Accelerations Ratios on Cora}
  \begin{tabular}{cccc}
    \toprule
    (2-layered) Model & Model Size & Data Size & Calculations \\
    \midrule
    GCN&  360K & 14.8M  & 249,954,739  \\
    Bi-GCN& 11.53K & 0.47M & 4,669,515 \\
    \midrule
    Ratio & 31.2x& 31.5x& 53.5x\\
    \bottomrule
  \end{tabular}
  \label{table1}
\end{table}

\subsection{Data Size Compression}
Currently, the loaded data tends to contribute the majority of the memory consumptions.
In the commonly employed datasets, the node features tend to contribute the majority of the loaded data.
Thus, a binarization of the loaded node features can largely reduce the memory consumptions when GNNs process the datasets.
Note that the data size of the node features is employed as an approximation of the entire loaded data size in this paper, because the edges in commonly processed attribute graphs are usually sparse and the size of the division mask is also small.

Let the loaded node features be denoted as $X\in\mathbb{R}^{N\times d}$, where $N$ is the number of nodes and $d$ is the number of features per node.
Then, the full-precision $X$ contains $N\times d$ floating-point values.
In our Bi-GCN, the loaded data $X$ can be binarized.
Then, $N\times d$ binary values and $N$ floating-point values can be obtained.
Thus, the size of the loaded data $X$ can be reduced by a factor of
\begin{equation}
  DC =\frac{32Nd}{Nd+32N} = \frac{32d}{d+32}.
  \label{DC}
\end{equation}

According to Eq. \ref{DC}, the compression ratio of the loaded data size depends on the dimension of the node features.
In practice, Bi-GCN can achieve an average reduction of memory consumption with a factor of $\thicksim$31x, which indicates that a much bigger attributed graph can be entirely loaded with identical memory consumption.
For some inductive datasets, we can then successfully load the entire graph or utilize a bigger sub-graph than that in the full-precision GCN.
The results of data size compression can be found in Tables \ref{table-trans} and \ref{table-ind}.


\subsection{Acceleration}
After the analysis of memory consumptions, the analysis of acceleration of our Bi-GCN, compared to GCN, is performed.
Let the input matrix and the parameters of the $l$-th layer possess the dimensions $N\times d_{in}^{(l)}$ and $d_{in}^{(l)}\times d_{out}^{(l)}$, respectively.
The original feature extraction step in GCN requires $Nd_{in}^{(l)}d_{out}^{(l)}$ addition and $Nd_{in}^{(l)}d_{out}^{(l)}$ multiplication operations. 
On the contrary, the binarized feature extraction step in our Bi-GCN only requires $Nd_{in}^{(l)}d_{out}^{(l)}$ binary operations and $2Nd_{out}$ floating-point multiplication operations. 
According to \cite{xnornet}, the processing time of performing one cycle operation, which contains one multiplication and one addition, can be utilized to perform 64 binary operations.
Then, the acceleration ratio for the feature extraction step of the $l$-th layer can be calculated as
\begin{equation}
  S^{(l)}_{fe} =\frac{Nd_{in}^{(l)}d_{out}^{(l)}}{\frac{1}{64}Nd_{in}^{(l)}d_{out}^{(l)}+2Nd_{out}^{(l)}} = \frac{64d_{in}^{(l)}}{d_{in}^{(l)}+128}.  
  \label{S-fe}
\end{equation}
As can be observed from Eq. \ref{S-fe}, the dimension of the node features $d_{in}^{(l)}$ determines the acceleration efficiency for the feature extraction step. 

For the aggregation step, the sparse matrix multiplication contains $|\mathcal{E}|d_{out}^{(l)}$ floating-point addition and $|\mathcal{E}|d_{out}^{(l)}$ floating-point multiplication operations.
If we let the average degree of the nodes be $\overline{deg}$, then $|\mathcal{E}| = N\overline{deg}/2$.

Therefore, the complete acceleration ratio of the $l$-th graph convolutional layer can be approximately computed via
\begin{equation}
  \begin{aligned}
    S^{(l)}_{full} &=\frac{Nd_{in}^{(l)}d_{out}^{(l)}+|\mathcal{E}|d_{out}^{(l)}}{\frac{1}{64}Nd_{in}^{(l)}d_{out}^{(l)}+2Nd_{out}^{(l)}+|\mathcal{E}|d_{out}^{(l)}} \\
    & = \frac{64d_{in}^{(l)}+32\overline{deg}}{d_{in}^{(l)}+128+32\overline{deg}}.
  \end{aligned}
  \label{S-full}
\end{equation}
Note that the average degree $\overline{deg}$ is usually small in the benchmark datasets, e.g., $\overline{deg}\approx 2.0$ in the Cora dataset.
When processing a graph with a low average node degree, the computational cost for the aggregation step, i.e., $32\overline{deg}$, usually possesses negligible effect on the acceleration ratio.
Thus, the acceleration ratio of the $l$-th layer can be approximately computed via
\begin{equation}
    S^{(l)}_{full} \approx S^{(l)}_{fe}.
    \label{S-app}
\end{equation}
Therefore, when $\overline{deg}$ is small, the acceleration ratio mainly depends on the input dimension of the binarized graph convolutional layers, according to Eqs. \ref{S-fe} and \ref{S-app}.
The input dimension of the first graph convolutional layer equals to the dimension of the node features in the input graph.
The input dimensions of the other graph convolutional layers equal to the dimensions of the hidden layers.
Since the dimension of the input node features is usually large, the acceleration ratio tends to be high for the first layer, e.g., $\thicksim$59x on the Cora dataset.
In general, the layer with a larger input dimension tends to require more calculations and can thus save more calculations with our binarization.
For example, the acceleration ratio of a 2-layered Bi-GCN on the Cora dataset can achieve $\thicksim$59x acceleration ratio for the first layer and $\thicksim$21x for the second layer.
In total, our 2-layered Bi-GCN can achieve $\thicksim$53x acceleration ratio on the Cora dataset.

\section{Capacity Exploration}
The proposed Bi-GNNs intend to extract the critical information by using a much smaller network than the floating-point GNNs.
Thus, they are more efficient in computational complexities and memory consumptions.
Unfortunately, similar to other simple networks, Bi-GNNs can easily be bothered by the \textit{capacity problem} when the storage capacity is less than demand in processing a certain task.
For example, if a floating-point GCN is well-trained with exactly enough parameters for a task, a direct binarization of the original structure may meet the \textit{capacity problem}, i.e., it may not possess enough representation ability to learn the task decently.
To theoretically tackle this problem, we propose a simple yet effective Entropy Cover Hypothesis which explores the lower bound of the proper width of binary hidden layers of Bi-GNNs to the floating-point GNNs for a particular task.
Note that we still take Bi-GCN as the example in this section.

\subsection{Entropy Cover Hypothesis}
Firstly, let us consider a 2-layered floating-point GCN in a node classification task, which is well-trained on a typical dataset, e.g., PubMed \cite{citation}.
Assume the distribution of each hidden neuron is accessible.
Let $d_{in}, d_{fp}, d_{out}$ be the dimensions of its input layer, hidden layer, and output layer, respectively, with $d_{in}\geq d_{fp} \geq d_{out}$.
Let $\mathcal{H}(x), \mathcal{H}(h), \mathcal{H}(\hat{y})$ denote the entropies of the input layer, hidden layer, and output layer, respectively.
Intuitively, the neural networks play a role as a semantic extractor, which gradually extracts the semantic-related information from the inputs.
Apparently, the total information is expected to be decreasing, i.e., $\mathcal{H}(x)\geq \mathcal{H}(h)\geq \mathcal{H}(\hat{y})$.

Then, if we directly binarize the above GCN, a straightforward 2-layered Bi-GCN can be obtained with the same input dimension $d_{in}$ and hidden dimension $d_{fp}$, and an approximately identical output $\hat{y}$, because the identical representation ability with GCN is expected.
Similarly, in Bi-GCN, the decreasing amount of information is also expected in ideal situation, i.e.,
\begin{equation}
  \mathcal{H}(x_b)\geq \mathcal{H}(h_b)\geq \mathcal{H}(\hat{y}),
\end{equation}
where $x_{b}$ represents the binary input feature and $h_{b}$ is the binary intermediate feature with $d_{fp}$ binary neurons.
However, since the expected output $\hat{y}$ is a floating-point output vector, $\mathcal{H}(h_b)\geq \mathcal{H}(\hat{y})$ is not always guaranteed.
If we directly utilize the binary hidden layers with the same dimension of the floating-point hidden layers in GCN, the \textit{capacity problem} may be induced.
Therefore, to achieve a comparable performance with the floating-point GCN, it is desired to explore the suitable number of binary hidden neurons, $d_{bin}$.

To tackle this problem, an intuitional assumption is that the storage capacity of binary hidden layer should be able to contain (no smaller than) the amount of information in the floating-point hidden layer, i.e., $\mathcal{H}(h)$.
The maximum entropy is utilized to represent the storage capacity \cite{entropy_and_generalization_in_feedforward_nets}, which is denoted as $C(h_b)$.
Then, the above assumption can be described as
\begin{equation}
  C(h_b) = d_{bin} \cdot \log_{2}2 = d_{bin} \geq \mathcal{H}(h).
\end{equation}
Under such circumstance, the number of binary hidden neurons should be no smaller than $\mathcal{H}(h)$.
Then, our Entropy Cover Hypothesis is defined as below.

\noindent\textbf{\textit{Entropy Cover Hypothesis:}} 

\textit{
  If a $L$-layered GCN generates the peak performance with $d_{fp}$ neurons in each hidden layer, the peak performance of a $L$-layered Bi-GCN can be obtained when the number of binary hidden neurons satisfies
  \begin{equation}
    d_{bin} \geq MAX\{\mathcal{H}(h^{(l)})|l=1,...,L-1\}.
  \end{equation}
  Specifically, with respect to a 2-layered GCN, the width of the binary hidden layers of a 2-layered Bi-GCN should satisfies
  \begin{equation}
    d_{bin} \geq \mathcal{H}(h^{(1)}).
  \end{equation}
}

The proposed Entropy Cover Hypothesis introduces a lower bound of the width of Bi-GCN hidden layers.
Note that its effectiveness will be verified experimentally in the latter section.
Then, the Entropy Cover Hypothesis can be utilized to guide the construction of the corresponding Bi-GCN by estimating the lower bound of the storage capacity.


\subsection{Entropy Estimation of Hidden Layers}
In this subsection, we will introduce the adopted estimation method for the entropies of the hidden layers.

Firstly, for a floating-point neuron in the hidden layer,
the \textit{binning} methods \cite{entropy_binning1,entropy_binning2} is utilized here to estimate its entropy, which divides the continuous space of a floating-point valued neuron into certain intervals and discretizes it into a finite number, $M$.
Then, given $N$ input samples, the empirical distribution of any floating-point valued neuron, e.g., $h_i$, can be formulated as 
\begin{equation}
  p_{h_i,m}\equiv\frac{1}{N}\sum_{j=1}^{N}\delta_m(h_{ij}),
\end{equation}
where $\delta_m$ denotes the probability measurement concentrated on $m$ \cite{estimation_of_entropy_and_mutual_information}.
Then, the entropy of a floating-point valued neuron can be estimated \cite{entropy_binning1,MLE-entropy}, i.e.,
\begin{equation}
  \hat{\mathcal{H}}(h_i)=-\sum_{m}^{M} p_{h_i,m} \log p_{h_i,m}.
\end{equation}

According to the chain rule of conditional probability, the joint entropy of a hidden layer $\mathcal{H}(h)=\mathcal{H}(h_1,h_2,...,h_{d_{fp}})$ can be decomposed as 
\begin{equation}
  \begin{aligned}
    \mathcal{H}\left(h\right)=&\mathcal{H}\left(h_1|h_2,\ldots,h_{d_{fp}}\right)+\\
    &\mathcal{H}\left(h_2|h_3,\ldots,h_{d_{fp}}\right)+...+\mathcal{H}\left(h_{d_{fp}}\right),
  \end{aligned}
  \label{entropy_of_layer}
\end{equation}
where $h_1, h_2, ..., h_{d_{fp}}$ are $d_{fp}$ neurons in the hidden layer $h$.
Eq. \ref{entropy_of_layer} is the exact calculation formula for the entropy of the hidden layer $\mathcal{H}\left(h\right)$.
However, it is difficult to directly compute $\mathcal{H}\left(h\right)$, due to the complicated dependencies among the hidden neurons.

Considering the situation that a GCN contains exactly enough parameters, the hidden representation of it may possess little redundancy.
Under such circumstance, the hidden neuron tends to be independent to each others to fully utilize the limited storage capacity.
Thus, 
\begin{equation}
  \mathcal{H}_{ind}(h) = \mathcal{H}(h_1)+\mathcal{H}(h_2)+...+\mathcal{H}(h_{d_{fp}})
\end{equation}
is exploited to estimate the joint entropy $\mathcal{H}(h)$, which can be calculated by the \textit{binning} \cite{MLE-entropy} method.
Then, the lower bound of the width of Bi-GNN hidden layers can be estimated according to our proposed Entropy Cover Hypothesis.


\section{Evaluations}
In this section, we evaluate the proposed binarization approach and our Bi-GNNs on benchmark datasets for the node classification task, and verify the effectiveness of our Entropy Cover Hypothesis.
Note that the memory consumptions and the number of cycle operations are theoretically estimated based on the specific settings of the methods and datasets.
Our codes are available at https://github.com/bywmm/Bi-GCN.

\begin{table}[t]
  \centering
  \caption{Datasets}
  \begin{tabular}{ccccc}
    \toprule
    Dataset & Nodes & Edges & Classes & Features\\
    \midrule
    Cora&  2,708 & 5,429 & 7 & 1,433\\
    CiteSeer& 3,327 & 4,732 & 6 & 3,703\\
    PubMed& 19,711 & 44,338 & 3 & 500\\
    OGBN-Arxiv & 169,343 & 1,166,243 & 40 &128\\
    \midrule
    Flickr & 89,250 & 899,756 & 7 & 500\\
    Reddit & 232,965 & 11,606,919 & 41 & 602\\
    OGBN-Products &2,449,029 &61,859,140 & 47 & 100\\
    \bottomrule
  \end{tabular}
  \label{table-datasets}
\end{table}

\subsection{Datasets}



We conduct our experiments with both the transductive and inductive learning settings on seven commonly employed datasets. 
The datasets are summarized in Table \ref{table-datasets}.

For the transductive learning task, four citation networks, i.e., Cora, PubMed, CiteSeer \cite{citation}, and OGBN-Arxiv \cite{ogb}, are utilized.
In these citation networks, nodes and edges represent the research papers (with the bag-of-words features) and citations (as undirected links), respectively. 
Articles are categorized into various classes according to the disciplines. 
We adopt the same data division strategy as OGB benchmarks \cite{ogb} for OGBN-Arxiv, and that as Planetoid \cite{planetoid} for Cora, PubMed, and CiteSeer.

Three datasets, i.e., Flickr, Reddit, and OGBN-Products, are employed for the inductive learning task.
Flickr is an image network collected by the SNAP website from four different sources.
Each node represents an image uploaded to Flickr with 500-dimensional bag-of-word features.
Undirected edges are formed between each pair of images captured at the same location, each pair of images sharing common tags, etc. 
We adopt the same data division strategy as GraphSAINT \cite{planetoid}.
Reddit is a post network constructed in \cite{graphsage}.
Each node represents a post with a GloVe 300-Dimensional Word Vector \cite{GloVe}.
A post-to-post connection is formed if the same user comments on both of the posts.
The label represents the community to which a post belongs.
The data division strategy in GraphSAGE \cite{graphsage} is adopted.
OGBN-Products \cite{ogb} is a co-purchasing network constructed on Amazon.
Nodes and edges represent products and the co-purchasing relationships between products, respectively.
Here, the node features are the bag-of-words features of the product descriptions \cite{cluster-gcn}.
The labels are the categories of the products.
We adopt the same data division and evaluation strategy as OGB benchmarks\cite{ogb}.

\begin{table*}[t]
  \footnotesize
  \centering
  \caption{Transductive learning results.(M.S., D.S., and C.O. are the abbreviations of Model Size, Data Size and Cycle Operations.)}
  \begin{tabular}{c|cccc|cccc|cccc}
    \toprule
    \multirow{2}{*}{Networks} & \multicolumn{4}{c|}{Cora} & \multicolumn{4}{c|}{PubMed} & \multicolumn{4}{c}{CiteSeer}\\
    \cline{2-13}
                 & Accuracy   & M.S.            & D.S.           & C.O.            & Accuracy    &  M.S.         & D.S.           & C.O.            & Accuracy    &  M.S.         & D.S.           & C.O.            \\
    \midrule
    GAT          & 83.0 ± 0.7 & 360.55K         & 14.8M          & 2.51e8          &  79.0 ± 0.3 &  126.27K      & 37.6M          & 6.44e8          & 72.5 ± 0.7  & 927.8K        & 47.0M          & 7.91e8          \\
    FastGCN      & 79.8 ± 0.3 & 360K            & 14.8M          & 2.50e8          &  79.1 ± 0.2 & 125.75K       & 37.6M          & 6.38e8          & 68.8 ± 0.6  & 927.25K       & 47.0M          & 7.90e8          \\
    SGC          & 81.0 ± 0.0 & 39.18K          & 14.8M          & 2.72e7          &  78.9 ± 0.0 & 5.86K         & 37.6M          & 2.98e7          & 71.9 ± 0.1  & 86.79K        & 47.0M          & 7.32e7          \\
    \midrule
    GCN          & \textbf{81.4 ± 0.4} & 360K   & 14.8M          & 2.50e8          &  79.0 ± 0.3 & 125.75K       & 37.6M          & 6.38e8          &\textbf{70.9 ± 0.5}& 927.25K & 47.0M          & 7.90e8          \\
    Bi-GCN-F     & 81.1 ± 0.4 & 360K            & \textbf{0.47M} & 2.50e8          &\textbf{79.4 ± 1.0}& 125.75K & \textbf{1.25M} & 6.38e8          & 69.5 ± 1.0  & 927.25K       & \textbf{1.48M} & 7.90e8          \\
    Bi-GCN-W     & 78.3 ± 1.5 & \textbf{11.53K} & 14.8M          & 2.50e8          &  75.5 ± 1.4 & \textbf{4.19K}& 37.6M          & 6.38e8          & 56.8 ± 1.7  &\textbf{29.25K}& 47.0M          & 7.90e8          \\
    Bi-GCN       & 81.2 ± 0.8 & \textbf{11.53K} & \textbf{0.47M} & \textbf{4.67e6} &  78.2 ± 1.0 & \textbf{4.19K}& \textbf{1.25M} & \textbf{1.55e7} & 68.8 ± 0.9  &\textbf{29.25K}& \textbf{1.48M} & \textbf{1.31e7} \\
    \bottomrule
  \end{tabular}
  \label{table-trans}
\end{table*}

\begin{table*}[t]
  \small
  \centering
  \caption{
    Inductive learning results. 
    (M.S., D.S., and C.O. are the abbreviations of Model Size, Data Size and Cycle Operations.)
  }
  \begin{tabular}{c|cccc|cccc}
    \toprule
    \multirow{2}{*}{Networks} & \multicolumn{4}{c|}{Reddit} & \multicolumn{4}{c}{Flickr}\\
    \cline{2-9}
    & F1-micro & M.S. & D.S. & C.O. & F1-micro & M.S. & D.S. & C.O.\\
    \midrule
    GCN & \textbf{93.8 ± 0.1} &643.00K& 534.99M & 4.18e10 &\textbf{50.9 ± 0.3} &507.00K&170.23M & 1.18e10\\
    Bi-GCN & 93.1 ± 0.2 & \textbf{21.25K}  & \textbf{17.61M}    & \textbf{4.18e9}& 50.2 ± 0.4&\textbf{16.87K}&\textbf{5.66M}&\textbf{4.65e8} \\
    \midrule
    GraphSAGE& 95.2 ± 0.1 & 1286.00K & 534.99M & 8.01e10 & \textbf{50.9 ± 1.0}&1014.00K &170.23M& 2.34e10 \\
    Bi-GraphSAGE & \textbf{95.3 ± 0.1} & \textbf{42.51K} & \textbf{17.61M} & \textbf{4.92e9} & 50.2 ± 0.4&\textbf{33.74K}& \textbf{5.66M}& \textbf{6.93e8}\\
    \midrule
    GraphSAINT& \textbf{95.9 ± 0.1} & 1798.00K& 534.99M  & 1.13e11&\textbf{52.1 ± 0.1}&1526.00K&170.23M&3.53e10\\
    Bi-GraphSAINT& 95.7 ± 0.1 & \textbf{139.62K} &\textbf{17.61M} & \textbf{1.04e10} & 50.8 ± 0.2&\textbf{65.25K}&\textbf{5.66M}&\textbf{1.28e9}\\
    \bottomrule
  \end{tabular}
  \label{table-ind}
\end{table*}

\subsection{Setups}
For three commonly used citation networks, i.e., Cora, PubMed, and CiteSeer, we select a 2-layered GCN \cite{gcn} with 64 neurons in the hidden layer as the baseline.
Our Bi-GCN is obtained by binarizing this GCN.
The evaluation protocol in \cite{gcn} is applied to Bi-GCN.
In the training process, GCN and Bi-GCN are both trained for a maximum of 1000 epochs with an early stopping condition at 100 epochs, by using the Adam \cite{adam} optimizer with a learning rate of 0.001.
The dropout layers are utilized in the training process with a dropout rate of 0.4, after binarizing the input of the intermediate layer.
We initialize the full-precision weights by Xavier initialization \cite{glorot}.
A standard batch normalization \cite{bn} (with zero mean and variance being one) is applied to the input feature vectors in Bi-GCN.
Note that we also investigate the influences of different model depths on classification performance. 
All the hyperparameters are set to be identical to the 2-layered case.

For Reddit and Flickr, we select an inductive version of GCN\cite{graphsage}, GraphSAGE \cite{graphsage}, and GraphSAINT \cite{graphsaint} as our baselines. 
Note that a 2-layered GraphSAINT model is employed for the fair comparisons.
The settings from their literatures are employed.
We will binarize all the feature extraction steps to generalize their corresponding binarized versions.
The hyper-parameters in our binarized models are set to be identical to their full-precision versions.

For OGBN-Products and OGBN-Arxiv, GCN, GraphSAGE, and GAT are selected as the baselines.
All of these GNNs are stacked with 3 layers.
We utilize 256 hidden neurons for GCN and GraphSAGE, 128 hidden neurons and 2 heads for GAT, on OGBN-Arxiv.
We set 512 hidden neurons for GCN and GraphSAGE, 256 hidden neurons and 2 heads for GAT, on OGBN-Products.
All the models are initialized by Xavier initialization \cite{glorot}.
They are trained with 500 epochs on OGBN-Arxiv and 20 epochs on OGBN-Products, by employing the Adam optimizer \cite{adam} with a learning rate of 0.001.
Dropout layers are utilized in the training process, with a dropout rate of 0.5 for the floating-point model.
Batch normalization layers are utilized after each floating-point GNN layer, while they are utilized before each Bi-GNN layer.
We adopt transductive learning settings for OGBN-Arxiv and inductive learning settings for OGBN-Products, which are commonly used in OGB benchmark evaluations.
In the training process on OGBN-Products, neighbor sampling strategy \cite{graphsage} is utilized to generate their inductive versions, where the sampled neighbor size is set to 20 for GCN and GraphSAGE, 10 for GAT.
The same hyper-parameters are adopted for their binarized versions.

\begin{table}[t]
  \small
  \centering
  \caption{Accuracy on two OGBN datasets. (D.S. and C.O. are the abbreviations of Data Size and Cycle Operations.)}
  \begin{tabular}{c|cccc}
    \toprule
    OGBN-Arxiv    & Accuracy            &  M.S.        & D.S.           & C.O.\\
    \midrule
    GCN           & \textbf{72.0 ± 0.3} & 424.00K      & 82.69M         & 1.90e10   \\
    Bi-GCN        & 69.7 ± 0.2          &\textbf{15.41K}&\textbf{3.23M} &\textbf{1.12e9}\\
    \midrule
    GraphSAGE          & \textbf{72.2 ± 0.2} &848.00K       & 82.69M         &3.74e10\\
    Bi-GraphSAGE       & 69.2 ± 0.3          &\textbf{30.81K}&\textbf{3.23M} &\textbf{1.59e9} \\
    \midrule
    GAT           & \textbf{72.1 ± 0.2} & 428.31K      & 82.69M         & 2.03e10 \\
    Bi-GAT        & 70.3 ± 0.2          &\textbf{19.72K}&\textbf{3.23M} & \textbf{2.42e9}\\
    \bottomrule
    \toprule
    OGBN-Products & Accuracy            & M.S.           & D.S.            & C.O.\\
    \midrule
    GCN           & 77.2 ± 0.3          & 1318.00K       & 934.23M         &8.93e11\\
    Bi-GCN        & \textbf{78.2 ± 0.5} &\textbf{45.37K} &\textbf{38.54M}  &\textbf{8.44e10}\\
    \midrule
    GraphSAGE          & \textbf{79.3 ± 0.2} & 2636.00K       & 934.23M         &1.72e12\\
    Bi-GraphSAGE       & 78.3 ± 0.2          &\textbf{90.74K} &\textbf{38.54M}  & \textbf{1.03e11}\\
    \midrule 
    GAT           & \textbf{79.6 ± 0.5} & 1326.37K       & 934.23M         &1.03e12\\
    Bi-GAT        & 78.6 ± 0.6          &\textbf{53.74K} & \textbf{38.54M} &\textbf{2.17e11}\\
    \bottomrule
  \end{tabular}
  \label{table-ogb}
\end{table}


\subsection{Results} 
\subsubsection{Comparisons}
The results on Cora, PubMed, and CiteSeer are shown in Table \ref{table-trans}.
As can be observed, our Bi-GCN gives a comparable performance compared to the full-precision GCN and other baselines.
Meanwhile, our Bi-GCN can achieve an average of $\thicksim$51x faster inference speed and $\thicksim$30x lower memory consumption than the vanilla GCN, FastGCN, and GAT, on three citation datasets.
Besides, the proposed Bi-GCN is more effective than SGC, especially in terms of the loaded data size. 
Note that the degradation of prediction accuracy on the CiteSeer dataset is worse than that on the other two datasets.
It may be induced by its smaller average node degrees, i.e., $\frac{|\mathcal{V}|}{|\mathcal{E}|}$, such that the node features appear to possess a larger portion of the total information in the input data.

Table \ref{table-ind} and Table \ref{table-ogb} show the results of our Bi-GNNs, i.e., Bi-GraphSAGE, Bi-GraphSAINT, and Bi-GAT on four benchmarks.
Similar to our Bi-GCN, our binarized GNNs can also significantly alleviate the memory consumptions of both the loaded data and model parameters, and reduce the number of calculations, with comparable performances.
The original data sizes of the Reddit and OGBN-Products datasets are 534.99M and 934.23M, respectively.
In comparison, our binarized GNNs only respectively demand 17.61M and 38.54M to load the data, which proves the significance of our binarization approach.
Note that the acceleration ratios of the binarized GNNs on the Reddit dataset are only $\thicksim$10x, because the average node degree is large, as discussed in Sec 5.3.
In general, these results prove that our binarization approach is efficient and can be successfully generalized to various GNNs.

\begin{figure}[t]
  \centering
  \subfigure[] { 
    \label{fig-layer-acc}
    \includegraphics[width=0.45\columnwidth]{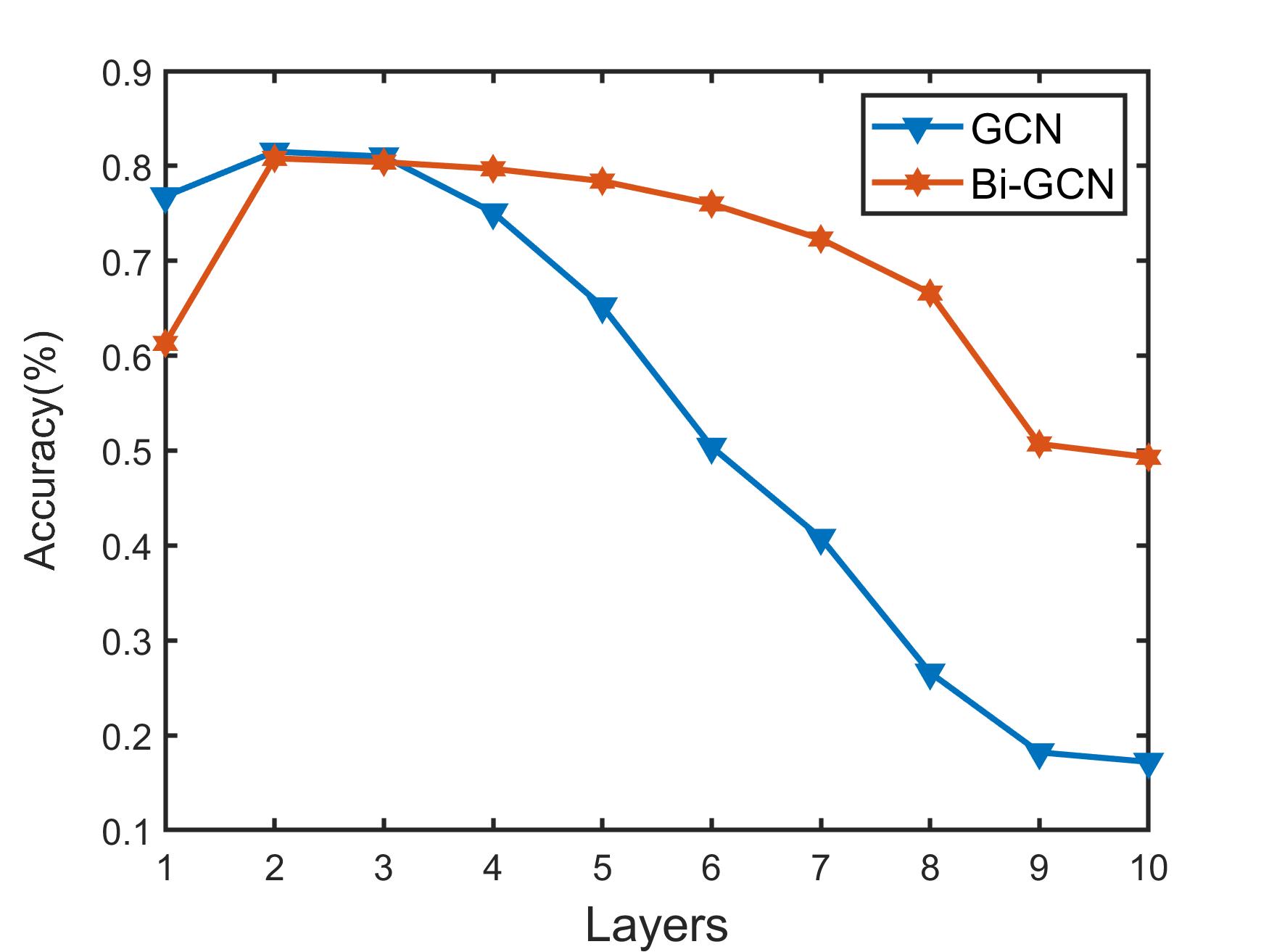}
  }
  \subfigure[] { 
    \label{fig-epoch-loss}
    \includegraphics[width=0.45\columnwidth]{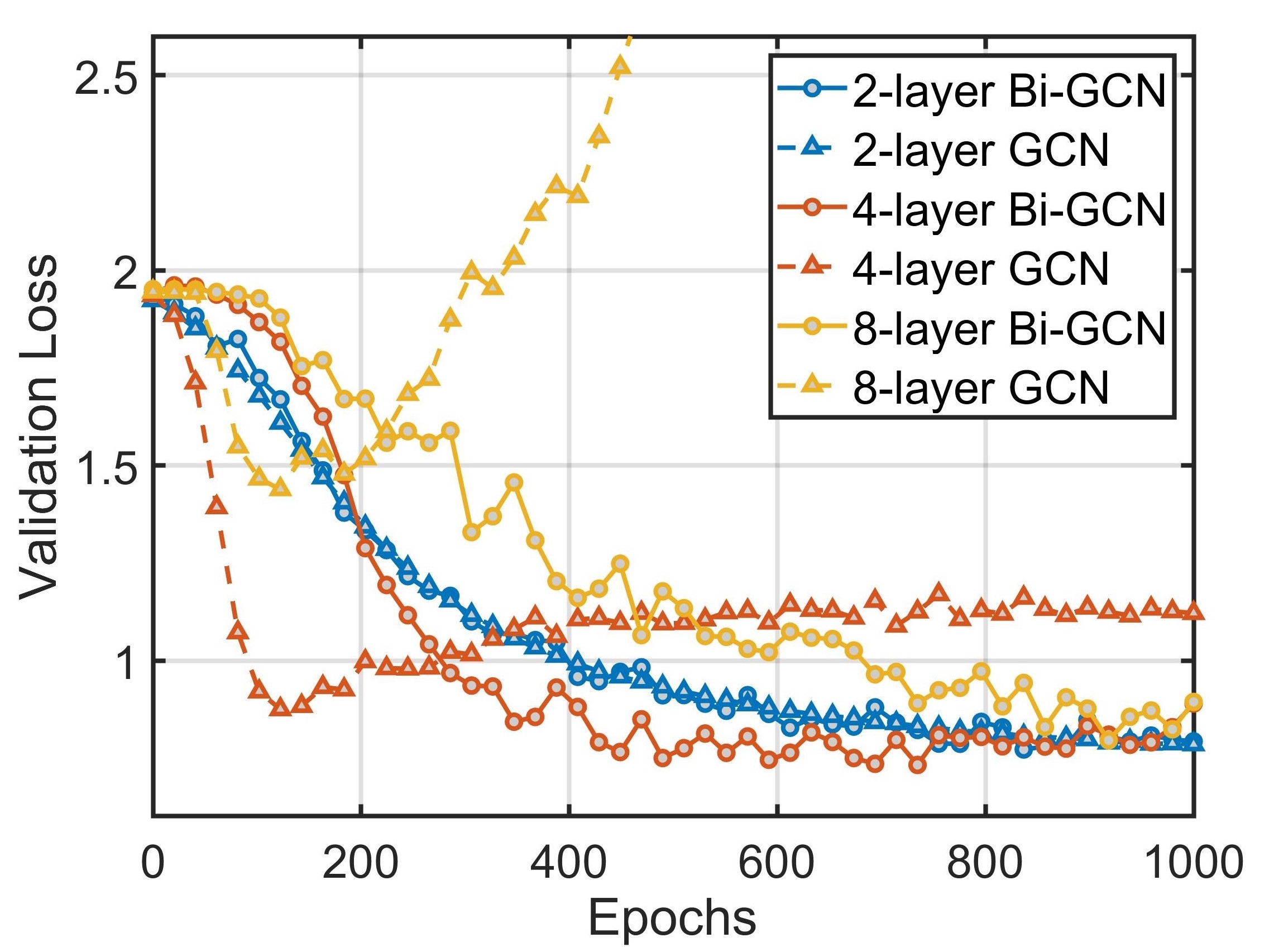}
  }
  \caption{Comparisons of accuracy and validation loss with different number of layers on the Cora dataset}
  \label{fig-layer-acc-and-epoch-loss}
\end{figure}

\begin{figure}[t]
  \centering
  \subfigure[] { 
    \label{fig-layer-memory}
    \includegraphics[width=0.45\columnwidth]{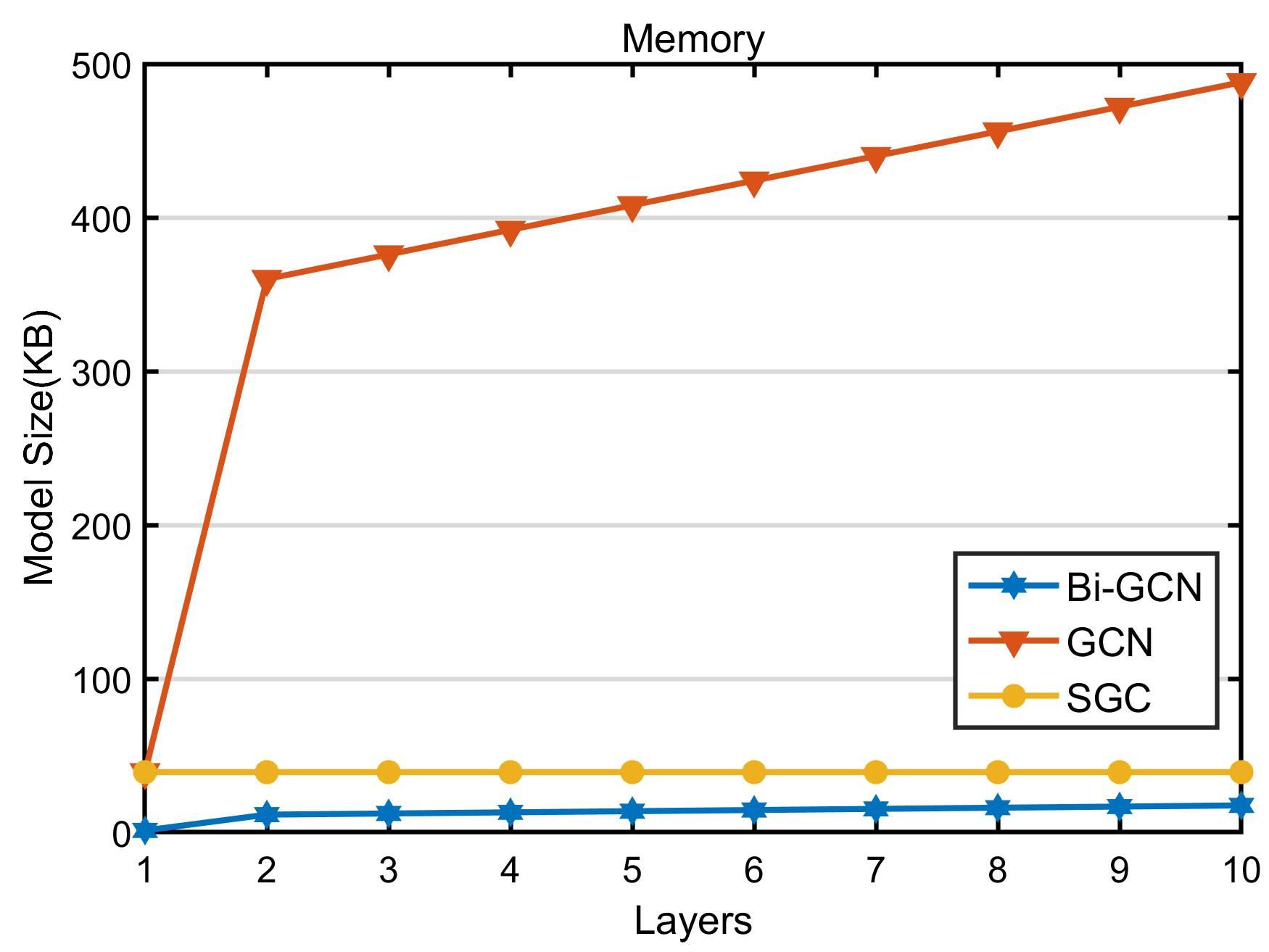}
  }
  \subfigure[] { 
    \label{fig5-layer-speed}
    \includegraphics[width=0.45\columnwidth]{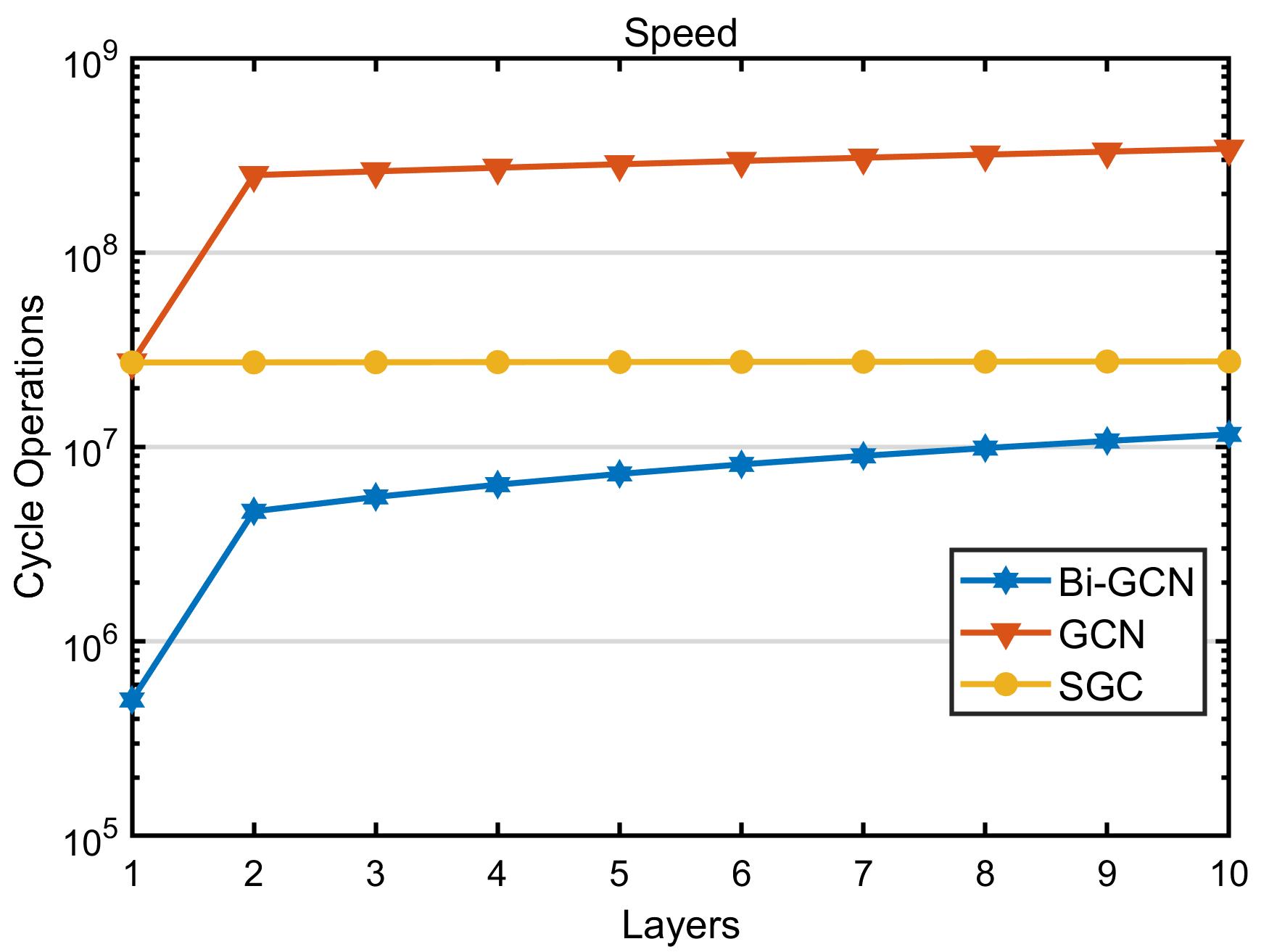}
  }
  \caption{Comparisons of memory consumption and inference speed on the Cora dataset.
  Note that SGC possesses only one layer and the x-axis of its results corresponds to the number of aggregations $K$.
  }
  \label{fig-layer-memory-speed}
\end{figure}

\begin{table}[t]
  \centering
  \caption{Entropy estimation of hidden layers in GNNs on three networks.
  Note that $n_h$* represents the number of neurons in the hidden layers.}
  \begin{tabular}{c|c|c|c|c}
    \toprule
    Datasets & GNN type($n_h$*) & num\_samples & $M$ & $\hat{H}_1(h)$\\
    \midrule
    PubMed & GCN(16)     & 19,711 & 200  & 97.37 \\
    \midrule
    Reddit & GCN (32)           & 232,965&1000  & 116.26  \\
    \midrule
    \multirow{2}{*}{OGBN-Products} 
    & GCN (128)          & 2,449,029&1000  & 414.67\\
    &GraphSAGE (512)          & 2,449,029&1000  & 849.29\\
    \bottomrule
  \end{tabular}
  \label{table-joint-entropy}
\end{table}

\subsubsection{Ablation Study}
Here, an ablation study is performed to verify the effectiveness of binarizing the network parameters and node features.
As can be observed from Table \ref{table-trans}, the prediction performances tend to vary less when the binarization is performed only to the node features.
This phenomenon indicates that there exists many redundancies in the full-precision features and our binarization can maintain the majority portion of effective information for node classification.
Meanwhile, the prediction results of binarizing the network parameters indicate that the binarized parameters cannot represent as much information as the full-precision parameters.
However, if both the node attributes and parameters are binarized, a comparable performance can be achieved, compared to GCN.
It reveals that the binarized network parameters can be effectively trained by the binarized features, i.e., Bi-GCN can successfully reduce the redundancies in the node representations, such that the useful cues can be learned well by a light-weighted binarized network.
Besides, binarizing the parameters and features separately can both reduce the memory consumptions, while the inference accelerations only appear when both the parameters and features are binarized.

\subsubsection{Effects of Different Model Depths}
Here, we analyze the effects of different model depths in our Bi-GCN.
Fig. \ref{fig-layer-acc} shows the transductive results of GCN and Bi-GCN on the Cora dataset with different model depths.
As can be observed, Bi-GCN is more suitable for constructing a deeper GNN than the original GCN. 
The accuracy of GCN has dropped sharply when it consists of three or more graph convolutional layers.
On the contrary, the performance of our Bi-GCN declines slowly.
According to Figure \ref{fig-epoch-loss}, GCN will quickly be bothered by the overfitting issue, as the number of layers increases.
However, our proposed Bi-GCN can effectively alleviate this overfitting problem.
Figure \ref{fig-layer-memory-speed} illustrates the comparisons of memory consumption and inference speed.
Since SGC contains only one layer, its memory consumption will not change with the increase of the number of aggregations.
When the number of layers increases, Bi-GCN can save more memories.
For the acceleration results, the ratio between GCN and Bi-GCN tends to decrease slightly when the number of layers increases, while the actual reduced computational costs increases.
Note that the required operations in SGC do not increase obviously because it only contains one feature extraction layer.

\begin{figure}[t]
  \centering
  \subfigure[Results on PubMed] { 
    \label{fig-neuron-acc-bigcn-pubmed}
    \includegraphics[width=0.45\columnwidth]{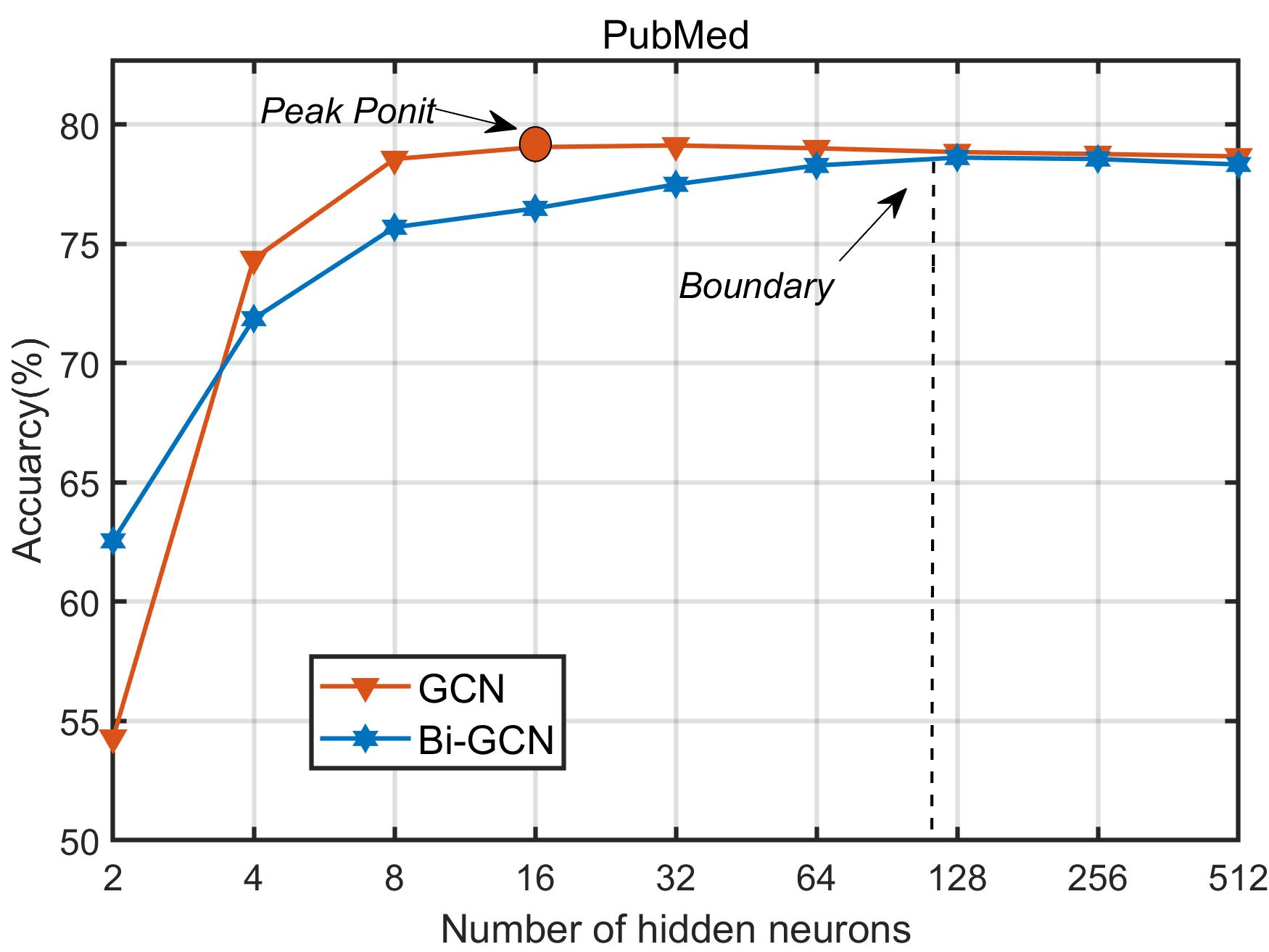}
  }
  \subfigure[Results on Reddit] { 
    \label{fig-neuron-acc-bigcn-reddit}
    \includegraphics[width=0.45\columnwidth]{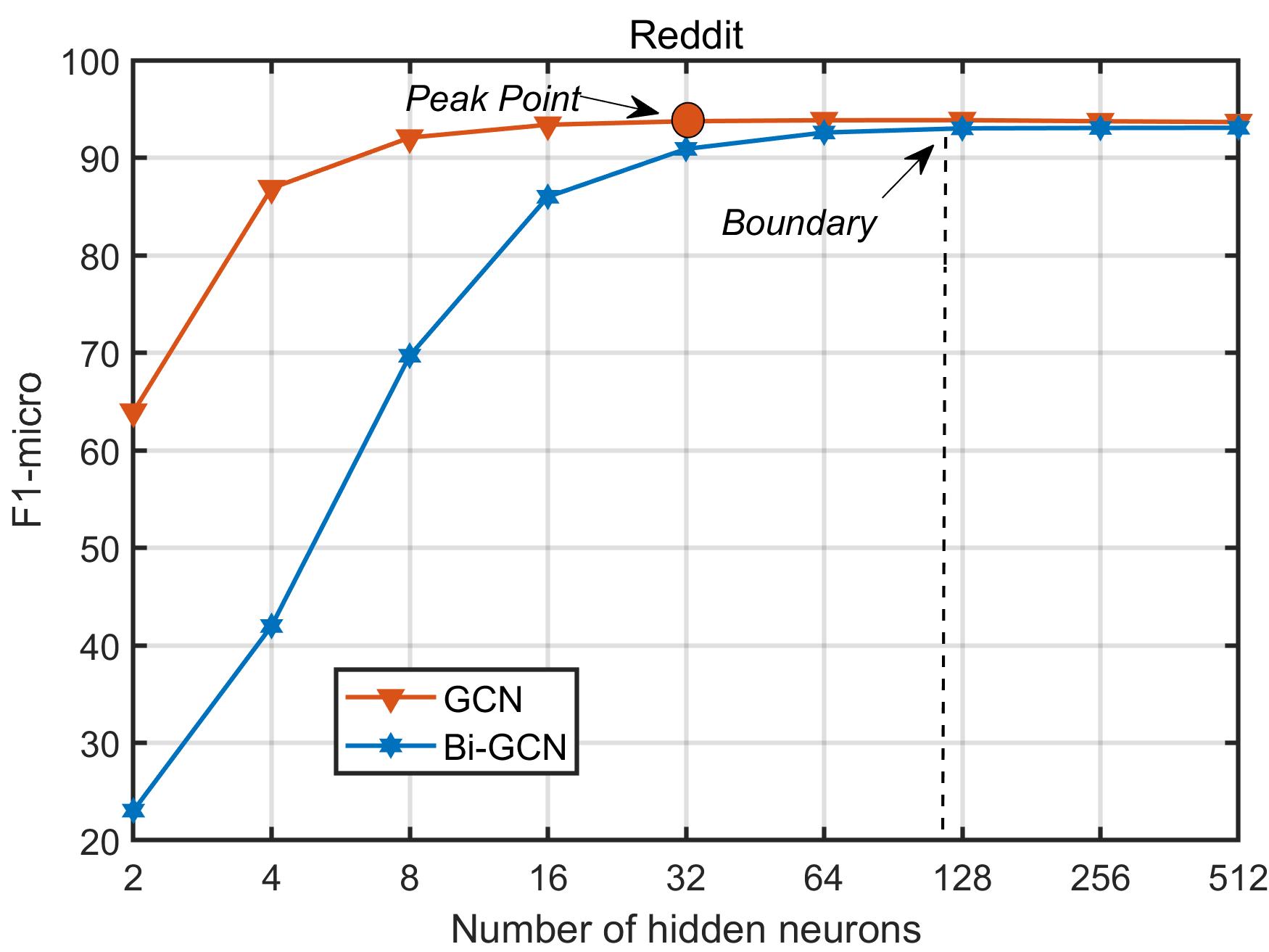}
  }
  \subfigure[Results on OGBN-Products] { 
    \label{fig-neuron-acc-bigcn-ogbnproducts}
    \includegraphics[width=0.45\columnwidth]{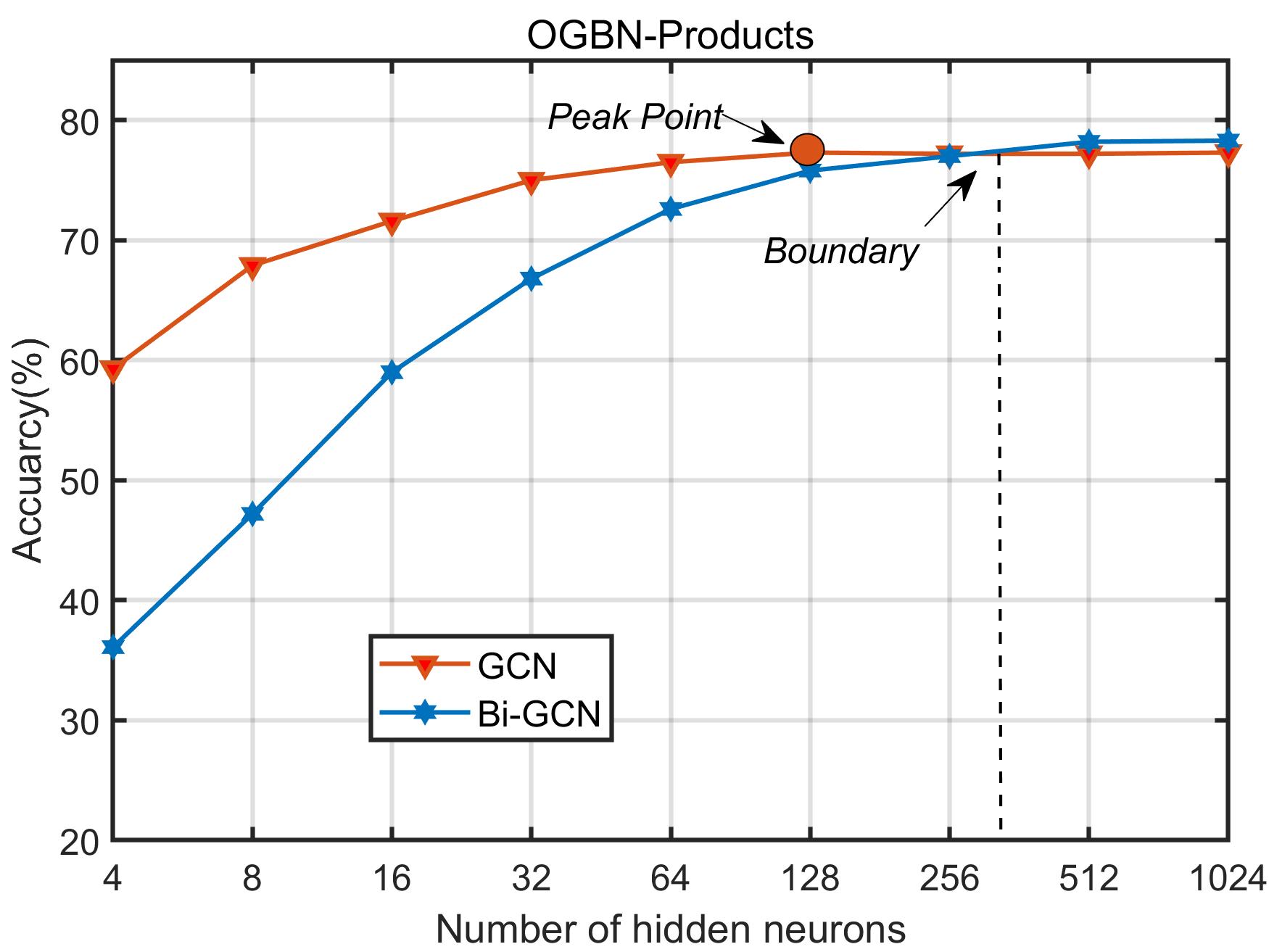}
  }
  \subfigure[Results on OGBN-Products] { 
    \label{fig-neuron-acc-bisage-ogbnproducts}
    \includegraphics[width=0.45\columnwidth]{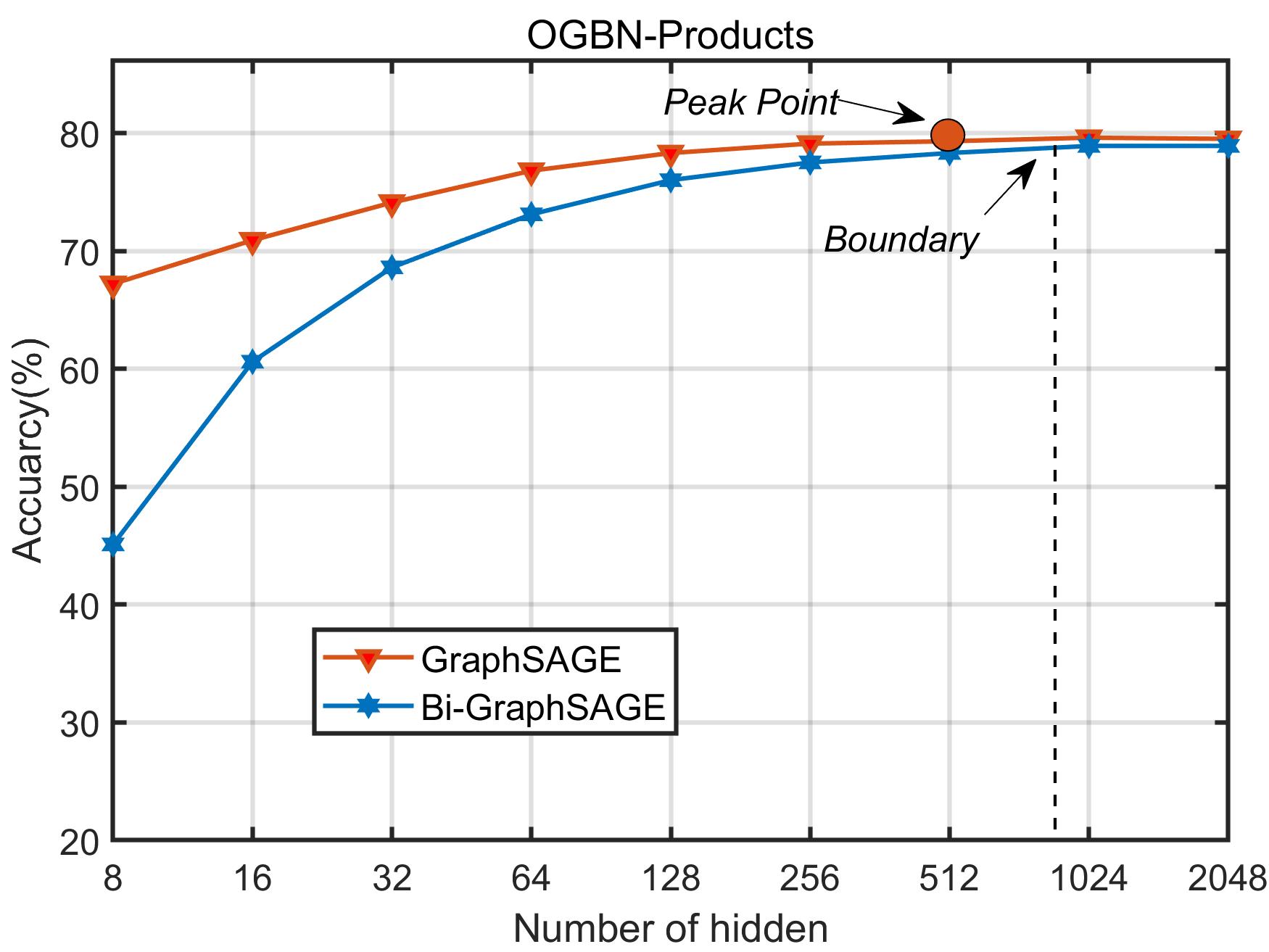}
  }
  \caption{The performances of different widths of the hidden layers for Bi-GNNs.}
  \label{fig-neuron-acc}
\end{figure}


\subsubsection{Capacity Analysis}
In this subsection, three node classification datasets with different scales and types, i.e., PubMed, Reddit, and OGBN-Products, are utilized to validate the effectiveness of the proposed Entropy Cover Hypothesis.
Fig. \ref{fig-neuron-acc} shows the results of GNNs and Bi-GNNs with different numbers of hidden neurons on three datasets.
As can be observed, the peak points, i.e., the minimum proper numbers of hidden neurons, are 16 for GCN on PubMed, 32 for GCN on Reddit, 128 for GCN on OGBN-Products, and 512 for GraphSAGE on OGBN-Products.
Note that the proper number of hidden neurons is the number, with which a (Bi-)GNN can achieve at least 99\% performance of its best performance.
According to our Entropy Cover Hypothesis, the boundary shows the lower bound of the width of the corresponding Bi-GNN hidden layers.
The detailed entropy estimation results are shown in Table \ref{table-joint-entropy}.
For example, 98 is the lower bound of the width of Bi-GCN hidden layer on the PubMed dataset, while the peak performance of Bi-GCN is achieved with 128 hidden neurons, which is slightly larger than the estimited lower bound.
Similar observations can be found in the other three cases.
According to the analysis in Sec. \ref{Sec-Analysis of Efficiency}, Bi-GCN can achieve $\thicksim$31x compression ratio for the model parameters, compared to floating-point GCN with the same number of hidden neurons.
Considering the extreme situation that both GCN and Bi-GCN achieve their peak performances with the minimum proper numbers of hidden neurons, Bi-GCN can also be $\thicksim$4x better on PubMed and $\thicksim$8x better on Reddit and OGBN-Products in terms of the model size. 
Besides, the compression efficiency of the loaded data does not vary with the changes of the width of hidden layers.
Since the size of typical data (graphs) is usually much larger than the model size, our Bi-GNN can still achieve $\thicksim$30x overall compression ratio on these three datasets.

\section{Conclusion}
This paper proposes a binarized version of GCN, named Bi-GCN, by binarizing the network parameters and the node attributes (input data).
The floating-point operations have been replaced by binary operations for inference acceleration.
Besides, we design a new gradient approximation based back-propagation method to train the binarized graph convolutional layers.
Based on our theoretical analysis, Bi-GCN can reduce the memory consumptions by an average of $\thicksim$31x for both the network parameters and node attributes, and accelerate the inference speed by an average of $\thicksim$51x, on three citation networks, i.e., Cora, PubMed, and CiteSeer.
Moreover, we introduce a general binarization approach applied to other GNNs, and the binarized GNNs (Bi-GNNs) can also obtain similar significant memory reductions and accelerations.
At last, an intuitive Entropy Cover Hypothesis is proposed to tackle the \textit{capacity problem} of Bi-GNNs by estimating the lower bound of the width of Bi-GNN hidden layers.
Extensive experiments have demonstrated that our Bi-GCN and Bi-GNNs can give comparable performance to the corresponding graph networks in both the transductive and inductive tasks and  verified the effectiveness of our Entropy Cover Hypothesis for solving the \textit{capacity problem}.


%






\ifCLASSOPTIONcaptionsoff
  \newpage
\fi


\bibliographystyle{IEEEtran}
\bibliography{IEEEtran.bib}
\end{document}